\documentclass[runningheads]{llncs}
\usepackage{graphicx}

\usepackage{amsmath,amssymb} 

\usepackage{comment}

\usepackage{xspace}
\newcommand*{\eg}{e.g.\@\xspace}
\newcommand*{\ie}{i.e.\@\xspace}
\newcommand*{\etal}{et al.\@\xspace}
\newcommand*{\etc}{etc.\@\xspace}
\usepackage{colortbl}
\usepackage{subfig}

\usepackage[pagebackref=true,breaklinks=true,colorlinks,bookmarks=false]{hyperref}
\usepackage{tabularx}
\usepackage{arydshln}
\usepackage{stmaryrd}
\usepackage{array,booktabs,arydshln,xcolor}
\usepackage{multirow}

\definecolor{shade2}{rgb}{1.,.89,.88}
\definecolor{shade4}{rgb}{.88,1.,1.}
\definecolor{shade3}{rgb}{1.,.94,.86}
\definecolor{shade1}{rgb}{0.9,0.9, 0.98}
\definecolor{shade5}{rgb}{1.,.89,.88}

\begin{document}

\pagestyle{headings}
\mainmatter

\def\ACCV22SubNumber{87}  

\title{Learning Generalizable Light Field Networks from Few Images} 
\titlerunning{Learning Generalizable Light Field Networks from Few Images}
\authorrunning{Qian Li, Franck Multon, Adnane Boukhayma}
\author{\large Qian Li \quad Franck Multon \quad Adnane Boukhayma}
\institute{Inria, Univ. Rennes, CNRS, IRISA, M2S, France}
\maketitle

\begin{abstract}
We explore a new strategy for few-shot novel view synthesis based on a neural light field representation. Given a target camera pose, an implicit neural network maps each ray to its target pixel's color directly. The network is conditioned on local ray features generated by coarse volumetric rendering from an explicit 3D feature volume. This volume is built from the input images using a 3D ConvNet. Our method achieves competitive performances on synthetic and real MVS data with respect to state-of-the-art neural radiance field based competition, while offering a $100$ times faster rendering. 
\end{abstract}

\begin{figure*}
\vspace{-10mm}
\centering
\includegraphics[width=\linewidth]{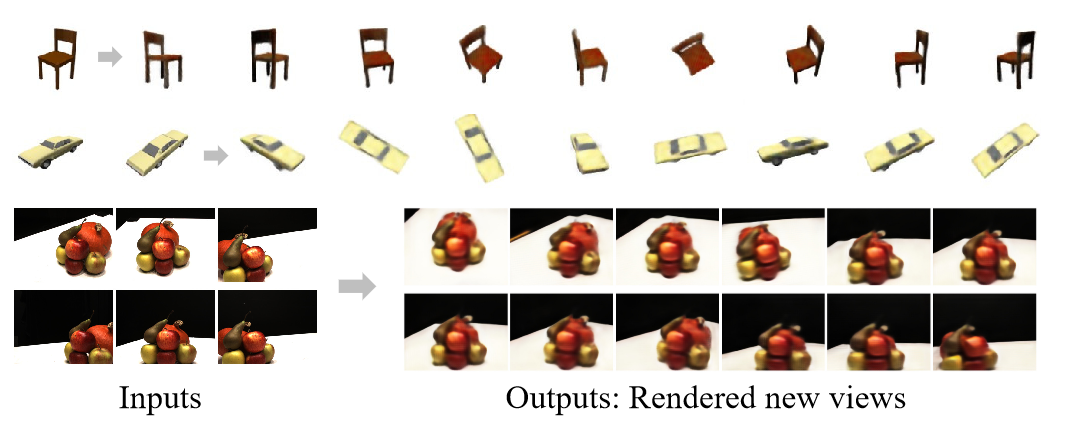}
  \caption{Our method enables fast generation of novel views from sparse input images without 3D supervision in training.}
  \label{fig:teaser}
  \vspace{-10mm}
\end{figure*}

\section{Introduction}

The ongoing research in computer vision and artificial intelligence has long sought to enable machines to understand and reason about the 3D world given limited observations. This ability is in fact crucial for many downstream 3D based machine learning, vision and graphics tasks.
Among these, novel view synthesis is a particularly prominent problem with numerous applications in free viewpoint and virtual reality, as well as image editing and manipulation. While traditional approaches required depth information, coarse geometric proxies or dense samplings of the input views, current deep learning based approaches are relying on deep neural network's generalization abilities across view points and 3D scenes to achieve novel view synthesis from minimal visual input.     

Leveraging a training data corpus of multi-view images and camera poses, without any 3D information, our goal is to build a novel view machine that can generalize outside the training data, and extrapolate beyond the input views. Given a few input calibrated color images at test time, we expect our method to generate novel target images given new query view points (See figure \ref{fig:teaser}). We are interested also in fast rendering novel view machines, that can predict new images in a single forward pass, without test time optimization.  
The recently popularized implicit neural representations offer numerous advantages in modelling 3D shape \cite{park2019deepsdf,mescheder2019occupancy} and appearance \cite{mildenhall2020nerf,sitzmann2019scene} in comparison to their traditional alternatives, while being conditionable using \eg encoders \cite{peng2020convolutional,chibane2020implicit,yu2021pixelnerf} and meta-learning \cite{sitzmann2020metasdf,ouasfi2022few}. In particular, Neural Radiance Fields \cite{mildenhall2020nerf} (NeRF) provide impressive novel view synthesis performances from dense input images. When coupled with convolutional encoders (e.g. \cite{yu2021pixelnerf,chen2021mvsnerf}), they can additionally achieve across-scene generalization and test-time optimization free inference, in addition to reconstructing from fewer inputs. However the rendering of these methods is expensive. In fact they require sampling hundreds of 3D points along each target pixel ray, evaluating densities and view-dependent colors for all these points through a multi-layer perceptron (MLP), and building the final image through volumetric rendering of all the samples' colors and densities. Multi-scale sampling is also necessary to achieve satisfactory results.

\begin{figure*}[t!]
\centering
\includegraphics[width=\linewidth]{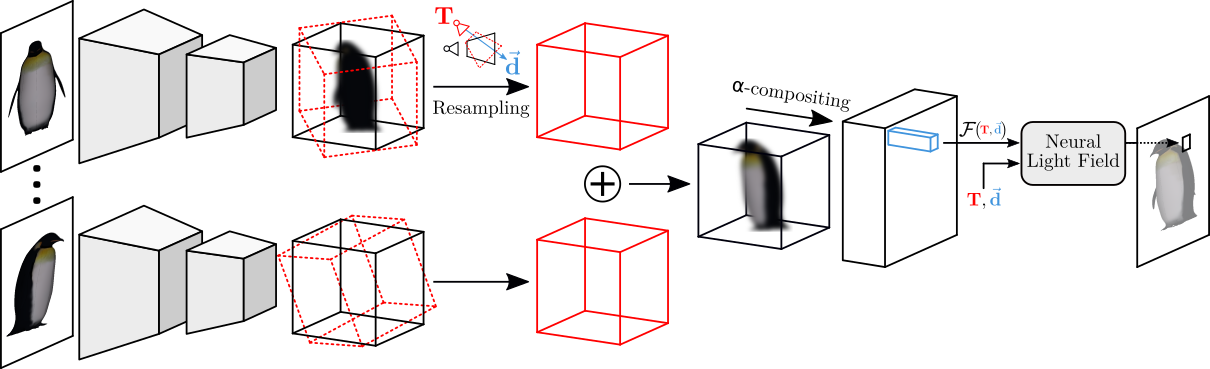}
  \caption{Overview of our method. Given an input image, a 3D feature volume is built with a convolutional neural network (first black cube). The volume represents features inside the input view frustum. Given a target view, these features are resampled into a volume representing the target view frustum (red cube). Target feature volumes originating from different input views are aggregated using learnable weights. An image of ray features is produced by rendering the target aggregated feature volume with $\alpha$-compositing. Finally the light field network maps a ray stemming from a target camera origin $T$ and spanning a direction $d$, along with its convolutional feature $\mathcal{F}$, to the corresponding pixel color of the target image.}
  \label{fig:pipe}
\end{figure*}

To reduce this complexity, we propose to use an implicit neural network operating in ray space rather than the 5D Euclidean $\times$ direction space, thus alleviating the need for per ray multi-sample evaluation and physical rendering. For a given target pixel, an MLP (\ie light field network) maps its ray coordinate and ray features to the color directly. Key to efficient generalization, and differently from \cite{sitzmann2021light}, we build the ray features by computing and merging 3D convolution feature volumes from the input images. These features are then rendered volumetrically into a coarse ray feature image, as illustrated in figure \ref{fig:pipe}. The method is fully differentiable and trained end-to-end.  

We evaluate our method on both synthetic (ShapeNet \cite{chang2015shapenet}) and real multi-view stereo data (DTU \cite{jensen2014large}). In the few-shot novel view optimization-free setting, we outperform comparable convolutional methods, including our own 3D convolutional baseline, and extend them to real complex data,
thanks to our hybrid approach combining 3D aware ConvNets and implicit neural representations. We achieve competitive results in comparison to generalizable encoder-decoder NeRF based models, while providing orders of magnitude faster rendering (see table \ref{tab:time}).    

\section{Related Work}
Novel view synthesis is  long-standing problem that spurred considerable work from the computer vision and graphics communities alike. While seminal work introduced various representations such as multi-plane images \cite{tucker2020single}, depth layered images \cite{shih20203d}, light fields \cite{mildenhall2019local}, depth based warping \cite{niklaus20193d,choi2019extreme} \etc, there has been a recent surge of particular interest in implicit neural shape and appearance representations (\eg \cite{mildenhall2020nerf,sitzmann2019scene,wang2021neus,kellnhofer2021neural,rematas2021sharf,wei2021nerfingmvs}) and neural rendering (\eg \cite{riegler2020free,riegler2021stable,thies2019deferred,aliev2020neural}). We discuss in this section existing work that we deemed most relevant to our contribution to few-shot novel view synthesis. 
 
Early deep learning based approaches to novel view from very sparse inputs used 2D convolutional encoder-decoder architectures mapping the input to the target image, conditioned on the target view. They either predicted colors directly \cite{tatarchenko2015single,Fworrall2017interpretable} or 2D flow fields \cite{zhou2016view,park2017transformation,sun2018multi} applied subsequently to the input. These methods were outperformed by 3D aware convolutional approaches, which rely on encoding then rendering explicit 3D latent presentations, either through volumetric rendering \cite{lombardi2019neural}, resterization \cite{wiles2020synsin}, or learnable neural rendering \cite{olszewski2019transformable,dupont2020equivariant}. The 3D latents take the form of intrinsic scene representations \cite{wiles2020synsin,hu2021worldsheet,chen2019monocular} or extrinsic volume grids \cite{olszewski2019transformable,dupont2020equivariant,lombardi2019neural}. 
Although many of these methods could learn to generate 360-degree views from very sparse inputs especially for synthetic central object data, most of them could not scale to high resolution images, complex scenes, and real data such as multi-view stereo datasets (DTU \cite{jensen2014large}). Some also required foreground segmentation masks at training (\eg \cite{olszewski2019transformable}). 

Implicit neural radiance fields (NeRF) \cite{mildenhall2020nerf} emerged later on as a powerful representation for novel view synthesis. It consists of an MLP mapping spatial points to volume density and view-dependant colors. Images are rendered through  hierarchical volumetric rendering. It presented initially however a few limitations such as computational and time rendering complexity, requiring dense training views, lacking across-scene generalization, and requiring test-time optimization. Current research work is tackling each of these limitations (\eg \cite{jain2021putting,niemeyer2021regnerf,barron2021mip,yu2021pixelnerf,liu2020neural,yu2021plenoctrees,garbin2021fastnerf,reiser2021kilonerf}). 
In particular, recent methods proposed to augment NeRFs with 2D \cite{yu2021pixelnerf,trevithick2021grf,wang2021ibrnet} and 3D \cite{chen2021mvsnerf} convolutional features collected from the input images. Hence, they offer forward pass prediction models, \ie test-time optimization free, while introducing generalization across scenes. However, they still need to evaluate hundreds of 3D query points per ray for inference similarly to NeRF, which makes them slow to render.
Furthermore, methods such as \cite{yu2021plenoctrees,hedman2021baking}
try to alleviate NeRFs' rendering complexity by learning view independent radiance features. In \cite{hedman2021baking}, the latter is combined with a single ray-dependant specular component, while Yu \etal \cite{yu2021plenoctrees} predict radiance spherical harmonic coefficients instead. Following \cite{sitzmann2021light}, we explore here a tangent strategy, consisting in bypassing 3D implicit radiance modelling all together.

Sitzmann \etal \cite{sitzmann2021light} introduced recently a new implicit representation for modelling multi-view appearance. A neural light field function maps rays \ie target pixels directly to their colors without any need for physical rendering.  However, this method was not demonstrated on real MVS data (\eg DTU \cite{jensen2014large}). It was also implemented in the auto-decoding setup, which means it requires test time optimization. The method also uses a hypernetwork for conditioning which makes it expensive to scale to bigger images in compute and memory. Differently, We propose here a more efficient local conditioning mechanism for the light field network, which allows us to scale to real data and larger images, and offers optimization-free inference.      

\section{Method}

Given one or few images $\{I_i\}$ of a scene or an object with their known camera parameters, \ie camera poses $\{R_i,T_i\}$, $R_i\in SO(3)$, $T_i\in\mathbb{R}^3$,  and intrinsics $K\in \mathbb{R}^{3\times 3}$ , our goal is to generate images $\{I_t\}$ for novel target views , \ie new camera poses $\{R_t,T_t\}$. We are interested in generalization to scenes and objects unseen at training, and target views beyond input view interpolation, using merely sparse input views. We also seek fast rendering time, and  we do not assume the availability of any segmentation masks neither at training nor testing. 

To this end, we propose a single forward pass inference deep learning method, that uses a deep neural network to map a ray $r$ in a projective pinhole camera model, to its desired color in the target view image $c_r$, using an implicit neural representation \ie a neural light field network $f$. This network is conditioned with ray features $\mathcal{F}_r$, \ie $c_r = f(r,\mathcal{F}_r)$. The ray features are generated through volumetric rendering of explicit 3D convolutional features built from the input images. A summary of our method is illustrated in figure \ref{fig:pipe}. We present in the remaining of this section the components of the two stages of our method, namely the convolutional stage, and the neural light field network.              
\subsection{Feature Volume}

Following seminal work (\eg  \cite{olszewski2019transformable,dupont2020equivariant,chen2021mvsnerf}), we build an explicit volume of features from an input image $I_i$ using a fully convolutional neural network $E$ consisting of a succession of a 2D convolutional U-Net and several 3D convolutional blocks:
\begin{align}
F_{i} = E(I_i),
\end{align}
where $I_i\in\mathbb{R}^{H\times W\times 3}$, $H$ and $W$ being the height and width of the input RGB image, and $F_i\in\mathbb{R}^{H_V\times W_V\times D\times C}$, $H_V$, $W_V$, $D$ and $C$ being respectively 
the height, width, depth, and the number of channels of the 3D feature volume. We refer the reader to the supplementary material for more details about the architecture of network $E$. The feature volume $F_i$ is expected to encode 3D shape and appearance information of the captured object or scene
in the view frustum associated with the input image, and is hence aligned pixel-wise with the latter. As we will show in the following sections, this volume will encode a prediction confidence, volume density \cite{mildenhall2020nerf}, colors, in addition to more generic appearance features. We note that one limitation of these features being modelled explicitly and not implicitly as in NeRF \cite{mildenhall2020nerf} based methods is that they cannot be view direction dependent. 

\subsection{Feature resampling} 

Next, using the the input feature volume $F_i$ aligned with the input image, we would like to create a feature volume $F_{t/i}$ aligned to the target image, that could be used subsequently to render a target feature image given the target camera pose $\{R_t,T_t\}$. Following the principles of volumetric rendering \cite{10.1145/800031.808594,mildenhall2020nerf}, in order to recreate a target image of dimensions $H_V\times W_V$, we need to evaluate $N$ points $\{p_{u,v}^z\}_{z=1}^{N}$ along each ray $r_{u,v}$ with direction $d_{u,v}$, where $u\in\llbracket 1,H_V\rrbracket$ and  $v\in\llbracket 1,W_V\rrbracket$: 
\begin{align}
d_{u,v} = R_tK^{-1}\begin{pmatrix} u\\ v\\ 1\\ \end{pmatrix}+T_t,
\quad
p_{u,v}^z = T_t + t_z\frac{d_{u,v}}{||d_{u,v}||},
\label{eq:d}
\end{align}

where $t_z\sim \mathcal{U}\left[z_n + \frac{z-1}{N}(z_f-z_n),z_n +\frac{z}{N}(z_f-z_n)\right]$ following \cite{mildenhall2020nerf}, $z_n$ and $z_f$ being the depth near and far bounds of the visual frustum. $K$ is the intrinsic camera matrix. The target volume $F_{t/i}$ is obtained then as the resampling of input volume $F_i$ with trilinear interpolation, using points $\{p_{u,v}^z\}$ aligned rigidly to the input camera coordinate frame: 
\begin{align}
F_{t/i}(u,v,z) = F_i(R_i^\top(p_{u,v}^z-T_i)),
\end{align}
where $F_{t/i}\in\mathbb{R}^{H_V\times W_V\times N\times C}$ and $\{R_i,T_i\}$ is the input camera pose. In practice, we normalize the aligned points' coordinates prior to sampling as $F_i$ is assumed to represent features in the input view normalized device coordinate (NDC) space. We use the NDC parametrization for an optimal spatial exploitation of the input feature volume $F_i$ and generalization across objects and scenes with different scales and datasets with different camera settings (\eg intrinsics, $z_n$, $z_f$).   

\subsection{Feature Aggregation}

As different input views provide different information about the observed scene, we merge subsequently the 3D features obtained from the various inputs. We note that all target feature volumes $\{F_{t/i}^k\}_k$ provided by input images $\{I_i^k\}_k$ are represented in the same target view camera coordinate frame. A naive merging strategy would be to simply average these volumes element-wise. However, for a given 3D location in the target view frustum, different input views contribute appearance information with varying confidence, based on the visibility/occlusion of this spatial location in the input views. In order to emulate this principle, and inspired by attention mechanisms, we propose to learn a 3D confidence measure per input view in the form of a weight volume $W_i\in\mathbb{R}^{H_V\times W_V\times D}$.
This volume is obtained as one of the channels of the input volume features $W_i = F_i(1)$ (\ie $W_{t/i} = F_{t/i}(1)$). As this confidence volume depends naturally on the input image and the relative camera pose of the target with respect to the input, similarly to \cite{sun2018multi}, we append these relative poses to the input image pixel values as additional input to the encoder $E$. After resampling the input features $\{F_i^k\}_k$ into the target ones $\{F_{t/i}^k\}_k$, we use the resampled weights $\{W_{t/i}^k\}_k$ normalized with Softmax across the input views to compute a weighted average of the target volumes: 
\begin{align} 
F_t = \sum_{k} \underset{k}{\text{Softmax}}(W_{t/i}^k) F_{t/i}^k(\rrbracket 1,C\rrbracket),
\end{align} 
where index $k$ is over the number of input views, and $F_{t}\in\mathbb{R}^{H_V\times W_V\times N\times C-1}$. Let us recall that this tensor represents features of $N$ points, within $[z_n,z_f]$ depth-wise, for all rays associated with a target image of dimension $H_V\times W_V$. This aggregation allows our method to use an arbitrary number of input views at both training and testing. In the single input case, we note that $F_t = F_{t/i}(\rrbracket 1,C\rrbracket)$.   

\subsection{Feature Rendering}

Following volumetric rendering \cite{10.1145/800031.808594,mildenhall2020nerf}, we generate a  target feature image $\tilde{\mathcal{F}}$ for a given target view differentiably using $\alpha$-compositing of the target feature volume $F_t$ along the depth dimension. To this end, we assume one of the target feature channels to represent volume density \cite{10.1145/800031.808594} $\sigma = F_t(1) \in \mathbb{R}^{H_V\times W_V\times D}$. We recall that the dimensions of tensor $F_t$ span the pixels of the target feature resolution $H_v\times W_v$ in the first two dimensions, and $N$ points sampled along each ray for the third dimension. The rendered target feature image then writes:
\begin{gather} 
\tilde{\mathcal{F}} = \sum_{z=1}^{N} T_{z}\alpha_zF_t(\rrbracket1,C-1\rrbracket),\\
T_z = e^{-\sum_{j=1}^{z-1}\sigma(j)\delta_j},
\quad \alpha_z = 1-e^{\sigma(z)\delta_z},
\label{eq:T}
\end{gather}
where $T$ represents transmittance,  $\delta_z = t_{z+1} - t_z$ and  $\tilde{\mathcal{F}}\in\mathbb{R}^{H_V\times W_V\times C-2}$.  In order to reduce the memory cost and increase the rendering speed of our method, the size of the rendered feature image is chosen to be lower than the size of the target image resolution, \ie $H_V = H/4$ and $W_V = W/4$. 

\subsection{Neural Light Field}

At this stage, the convotulional rendered features produce a low resolution feature image representative of all rays making up the target view. We propose to learn a light field function $f$, which performs both upsampling and refinement of the result of the convolutional first stage of our method. This implicit neural network maps rays of the target image to their colors, while being conditioned on ray features extracted from the convolutional rendered features. 

Given a ray $r_{u,v}$ with direction $d_{u,v}$ corresponding to the target image pixel coordinates $(u,v)$, with $(u,v)\in\llbracket1,H\rrbracket \times \llbracket1,W\rrbracket$, we encode rays using  Plücker coordinates similarly to Sitzmann \etal \cite{sitzmann2021light}: 
\begin{align}
r_{u,v} = \frac{(d_{u,v},T_t\times d_{u,v})}{||d_{u,v}||},
\end{align}
where $r_{u,v}\in\mathbb{R}^6$. This representation ensures a unique ray encoding when the origin $T_t$ moves along direction $d_{u,v}$. We recall that the expression of $d_{u,v}$ as a function of the target camera pose $\{R_t,T_t\}$ can be found in equation \ref{eq:d}.

The feature $\mathcal{F}_{u,v}$ of a ray $r_{u,v}$ at the final image resolution $H\times W$ is obtained from the lower resolution rendered feature image $\tilde{\mathcal{F}}\in\mathbb{R}^{H_V\times W_V\times C-2}$ through a learned upsampling. Specifically, the rendered feature image undergoes two successive 2D convolutions and up upsamplings to produce a feature image at the desired resolution $\mathcal{F}\in\mathbb{R}^{W\times H\times C-2}$. The final target RGB image $I_t=\{c_{u,v}\}_{u\in\llbracket1,H\rrbracket, v\in\llbracket1,W\rrbracket}$ is predicted then from the concatenation of the ray coordinate and its feature with an MLP accordingly:
\begin{align}
c_{u,v} = f(r_{u,v},\mathcal{F}_{u,v}). 
\end{align}

We refer the reader to the supplementary material for more details about the architecture of network $f$. Notice that while 
convolution equipped NeRF \cite{mildenhall2020nerf} methods (\eg pixelNeRF \cite{yu2021pixelnerf} MVSNeRF \cite{chen2021mvsnerf} GRF\cite{trevithick2021grf}) require querying $H\times W\times N$ 3D points through their implicit neural radiance fields, our light field network only needs to evaluate $H \times W$ rays, which enables our method to train potentially faster, and render orders of magnitude faster compared to \eg pixelNeRF (see Table \ref{tab:time}).

\subsection{Training Objective}

Our model is fully differentiable and trained end-to-end. We optimize the parameters of the model, namely the convolutional network $E$ and the light field network $f$ jointly, by back-propagating a combination of a fine loss $L_r$ and two coarse losses $\tilde{L}_r$ and $\tilde{L}_d$:
\begin{align}
L = L_r + \tilde{L}_r + \tilde{L}_d.
\end{align}
$L_r$ is a L2 reconstruction loss between the final image $I_t$ predicted by the light field network and the ground-truth $I_t^{gt}$:
\begin{align}
L_r = ||I_t - I_t^{gt}||^2_2.
\end{align}

We regularize the first convolutional stage of our method through a coarse L2 reconstruction loss $\tilde{L}_r$. After rendering the low resolution image feature $\tilde{\mathcal{F}}$, we constrain its first 3 channels to produce the RGB colors of the target ground truth image:
\begin{align}
\tilde{L}_r = ||\tilde{I}_t - \tilde{I}_t^{gt}||^2_2,
\end{align}
where $\tilde{I}_t = \tilde{\mathcal{F}}(\llbracket 1,3 \rrbracket)$, and $\tilde{I}_t^{gt}$ is the ground truth image downsampled to resolution $H_V\times W_V$. 

Additionally, we regularize the gradient of the low resolution depth image $\tilde{d}_t$ rendered from the density volume $\sigma$ of the first stage. Similarly to \cite{heise2013pm}, we weight this cost with an edge-aware term using the ground-truth image gradient:
\begin{align}
\tilde{L}_d = \tfrac{1}{H_V\times W_V} \sum_{u,v}{|\partial_u\tilde{d}_t|e^{-||\partial_u\tilde{I}_t^{gt}||}
+ 
|\partial_v\tilde{d}_t|e^{-||\partial_v\tilde{I}_t^{gt}||}}.
\end{align}
The depth image $\tilde{d}_t$ can be expressed as a function of transmittance $T$ and $\alpha$ values as follows:
\begin{align}
\tilde{d}_t = \tfrac{1}{\sum_{z=1}^{N}T_z\alpha_z}
\sum_{z=1}^{N}T_z\alpha_zt_z.
\end{align}
We note that the expressions of $T$ and $\alpha$ are detailed in equation \ref{eq:T}.

\section{Experiments}

We demonstrate the capability of our method to generate novel views from sparse input views using standard benchmarks. Specifically, we evaluate the ability of our network to reconstruct 360-degree novel views of objects unseen at training in the synthetic ShapeNet dataset \cite{sitzmann2019scene}. We additionally evaluate our ability to infer novel views of real world scenes using the more challenging DTU multi-view stereo dataset \cite{jensen2014large}.      


\subsection{Implementation Details}
We implemented our method with the PyTorch \cite{paszke2019pytorch} framework on a Quadro RTX 5000 gpu. We optimize with the Adam \cite{kingma2014adam} solver using learning rate $10^{-4}$ in training and $10^{-5}$ in fine-tuning. The depth of the convolutional feature volume is set to $D = 32$, and the number of channels $C=32$. For the MLP of the light field network, we use 5 layers with a hidden dimension of 256, similarly to NeRF \cite{mildenhall2020nerf}. For volumetric feature resampling, we use the coarse and fine sampling strategy similarly to previous work (\eg\cite{mildenhall2020nerf,yu2021pixelnerf}), with $N=64$ coarse samples and $N=32$ fine samples. We show detailed network structure in the supplementary material.

\subsection{Generalization on Synthetic Data}

\textbf{Dataset} Following the settings in GRF \cite{trevithick2021grf} and PixelNeRF\cite{yu2021pixelnerf}, we evaluate first our synthesis results on the car and chair classes of dataset ShapeNet-V2\cite{sitzmann2019scene}. Precisely, the cars amount to 2151 training objects, 352 validation objects and 704 testing objects, while the chairs count 4612 training objects, 662 validation objects and 1317 testing objects. Hence the testing objects are not seen in training. In the training split, each object has 50 images with size $128 \times 128$. For testing, there are 251 views per object. Our model takes 1 or 2 fixed views as input and infers novel views for evaluation.\\

\begin{table}
 \vspace{-6mm}
\begin{center}
\begin{tabular}{l|cc|cc|cc|cc}
\toprule[1.2pt]
Method & \multicolumn{2}{c} {\bf PSNR(Cars)$\uparrow$} &\multicolumn{2}{|c|}{\bf SSIM(Cars)$\uparrow$}& \multicolumn{2}{c}{\bf PSNR(Chairs)$\uparrow$} &\multicolumn{2}{|c}{\bf SSIM(Chairs)$\uparrow$} \\
   & 1-view & 2-view & 1-view & 2-view & 1-view & 2-view  & 1-view & 2-view   \\
\hline
SRN\cite{sitzmann2019scene} & 20.72 & 22.94  & 0.85 & 0.88 &22.89 &  24.48 & 0.91 & 0.92\\
LFN\cite{sitzmann2021light} & 22.42 & -- & 0.89 & -- & 22.26 & -- & 0.90 & -- \\
\hline
TCO\cite{tatarchenko2015single} & 18.15 & 18.41  & 0.79  & 0.80 & 21.27 & 21.33 & \cellcolor{shade1}0.88 & 0.88\\
WRL\cite{Fworrall2017interpretable} & 16.89 & 17.20 & 0.77 & 0.78 & 22.11 & 22.28 &\cellcolor{shade3}  0.90 & \cellcolor{shade1} 0.90\\
dGQN\cite{eslami2018neural} & 18.19 & 18.79  & 0.78 & 0.79  & 21.59 & 22.36 & 0.87 & 0.89\\
PixelNeRF\cite{yu2021pixelnerf} & \cellcolor{shade5} 23.17 &  \cellcolor{shade5} 25.66  &  \cellcolor{shade5} 0.90 &  \cellcolor{shade5} 0.94 &  \cellcolor{shade5} 23.72 & \cellcolor{shade5}  26.20 &  \cellcolor{shade5} 0.91 & \cellcolor{shade5}  0.94\\
GRF\cite{trevithick2021grf} & 20.33 & \cellcolor{shade1} 22.34  &  \cellcolor{shade1}0.82 & \cellcolor{shade1} 0.86  & 21.25 & \cellcolor{shade1} 22.65 & 0.86 & 0.88\\
ENR\cite{dupont2020equivariant} & \cellcolor{shade1} 22.26 & -- & -- & -- &\cellcolor{shade3}  22.83 & -- & -- & -- \\
{\bf Ours} & \cellcolor{shade3}  22.31 & \cellcolor{shade3}  23.82  & \cellcolor{shade3} 0.87 &\cellcolor{shade3}   0.91  & \cellcolor{shade1} 22.52 &\cellcolor{shade3}  24.10 & \cellcolor{shade3} 0.90 & \cellcolor{shade3}  0.92\\
\bottomrule[1.2pt]
\end{tabular}
\caption{Comparison of the average PSNR and SSIM of reconstructed images in the ShapeNet-V2 \cite{sitzmann2019scene} dataset. The higher the better for both PSNR and SSIM. Red, orange and blue represent the first, second and third ranking methods respectively.
SRN\cite{sitzmann2019scene} and LFN\cite{sitzmann2021light} require test time optimization.}
\label{tab:shapenet}
\end{center}
 \vspace{-8mm}
\end{table}

\begin{table}
 \vspace{-10mm}
\begin{center}
\label{table:headings}
\begin{tabular}{l|c|c}
\toprule[1.2pt]
 & {pixelNeRF\cite{yu2021pixelnerf}} & {\textbf{ours}}\\
\hline
clock time for a $128\times 128$ image in seconds & 6.23 & \textbf{0.06} \\
\bottomrule[1.2pt]
\end{tabular}
\caption{Comparison of rendering complexity. All clock times are collected on a Quadro RTX 5000 gpu. Our rendering is 100 times faster than PixelNeRF \cite{yu2021pixelnerf}.}
\label{tab:time}
\end{center}
 \vspace{-10mm}
\end{table}

\begin{figure}
\centering
\def\tabularxcolumn#1{m{#1}}
\begin{tabularx}{\linewidth}{@{}X@{}}
\setlength{\tabcolsep}{-1pt}
\begin{tabular}{ccccc | ccccc}
\multicolumn{5} {c|}{\bf 1 input view} &  \multicolumn{5}{c}{\bf 2 input views}\\
\includegraphics[width=1.2cm]{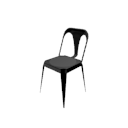} &
\includegraphics[width=1.2cm]{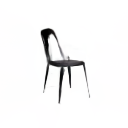} & \includegraphics[width=1.2cm]{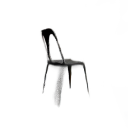} &
\includegraphics[width=1.2cm]{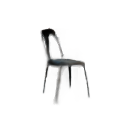} & \includegraphics[width=1.2cm]{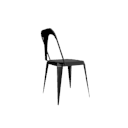} &
\includegraphics[width=1.2cm]{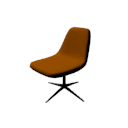}&
\includegraphics[width=1.2cm]{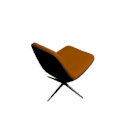} & \includegraphics[width=1.2cm]{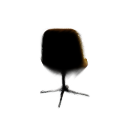}& \includegraphics[width=1.2cm]{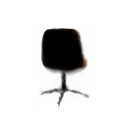} & \includegraphics[width=1.2cm]{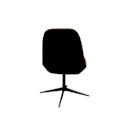} \\
\includegraphics[width=1.2cm]{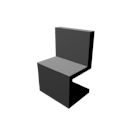} & \includegraphics[width=1.2cm]{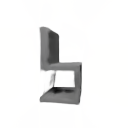} & \includegraphics[width=1.2cm]{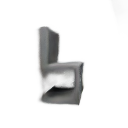} & \includegraphics[width=1.2cm]{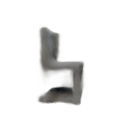} & \includegraphics[width=1.2cm]{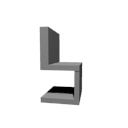} & \includegraphics[width=1.2cm]{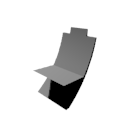} & \includegraphics[width=1.2cm]{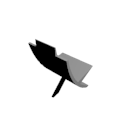} & \includegraphics[width=1.2cm]{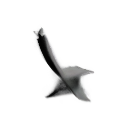} &
\includegraphics[width=1.2cm]{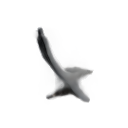} & \includegraphics[width=1.2cm]{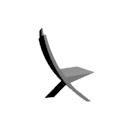} \\
\includegraphics[width=1.2cm]{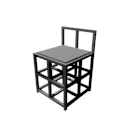} & \includegraphics[width=1.2cm]{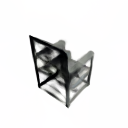} & \includegraphics[width=1.2cm]{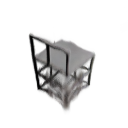} & \includegraphics[width=1.2cm]{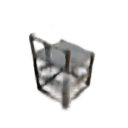} & \includegraphics[width=1.2cm]{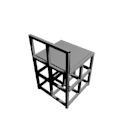} & \includegraphics[width=1.2cm]{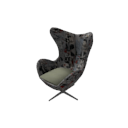} & \includegraphics[width=1.2cm]{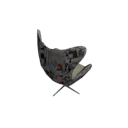} & \includegraphics[width=1.2cm]{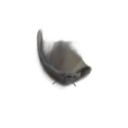} &
\includegraphics[width=1.2cm]{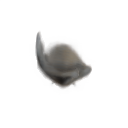} & \includegraphics[width=1.2cm]{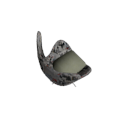} \\
Input & LFN & PixelNeRF & Ours   & GT&  \multicolumn{2}{c}{ Inputs} & PixelNeRF & Ours  & GT
\end{tabular}
\end{tabularx}
\vspace{-2mm}
 \caption{Qualitative comparison of novel view synthesis of unseen chairs from a single and 2 input views on ShapeNet-V2 \cite{sitzmann2019scene} (More results in \textbf{supplementary material}). LFN \cite{sitzmann2021light} requires test times optimization.} 
\label{fig:shapenet}
\vspace{-4mm}
\end{figure}

Table \ref{tab:shapenet} shows a quantitative comparison of our method with the recent state-of-the-art in few-shot view synthesis. We report the peak signal-to-noise ratio (PSNR) and structural similarity (SSIM) reconstruction metrics. We relay the numbers for methods TCO\cite{tatarchenko2015single}, WRL\cite{Fworrall2017interpretable}, dGQN\cite{eslami2018neural}
SRN\cite{sitzmann2019scene} and GRF \cite{trevithick2021grf} as they were reported in \cite{trevithick2021grf}. We report the numbers of ENR\cite{dupont2020equivariant} and PixelNeRF\cite{yu2021pixelnerf} from \cite{yu2021pixelnerf}, and the numbers for LFN\cite{sitzmann2021light} from their paper. Figure \ref{fig:shapenet} shows a qualitative comparison to a few of these methods. We obtain the visualizations for PixelNeRF\cite{yu2021pixelnerf} and LFN\cite{sitzmann2021light} using their publicly available codes and models. The teaser figure \ref{fig:teaser} shows additional 360-degree novel view synthesis results, and we provide further visual results in the supplementary material. 

The results confirm that 2D based image to image novel view methods (\eg TCO) are outperformed by the 3D aware ones (\eg ENR, SRN, etc).  Furthermore, 3D aware methods that use implicit 3D representations (\eg PixelNeRF, GRF, SRN) outperform generally their counterparts relying on explicit 3D latents (\eg ENR). Our method is hybrid, in that it uses an explicit 3D latent, combined with a 2D implicit representation.  

Numerically, our method is overall second to PixelNeRF, while being two orders of magnitude faster at rendering. As we show in table \ref{tab:time}, our method requires only $0.06$ seconds to infer a full $128\times128$ image, while PixelNeRF takes $6.23$ seconds using the implementation provided by the authors. As shown in figure \ref{fig:shapenet}, our method provides comparable reconstruction quality as well, while encountering similar difficulties for challenging views. 

While our performance is generally close to LFN across the benchmark, we note that LFN requires auto-decoding test time optimization. It also requires training hypernetworks, which are prohibitively expensive in compute and memory, thereby limiting the resolution of the reconstructed images. This hinders in turn the applicability of this method to real datasets with images larger than $128\times 128$. Conversely, we demonstrate the ability of our method to model complex real scenes with bigger images using moderate computational resources, while providing optimization-free single forward pass prediction. We note that SRN requires test-time optimization as well. 
\begin{figure}
\centering
\def\tabularxcolumn#1{m{#1}}
\begin{tabularx}{\linewidth}{@{}X@{}}
\setlength{\tabcolsep}{-1pt}
\begin{tabular}{cccc | ccccc}
\multicolumn{4} {c|}{\bf 1 input view} &  \multicolumn{5}{c}{\bf 2 input views}\\
 \includegraphics[width=1.4cm]{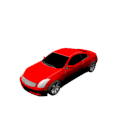} &  \includegraphics[width=1.4cm]{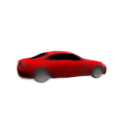} & \includegraphics[width=1.4cm]{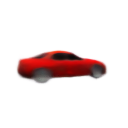}&
 \includegraphics[width=1.4cm]{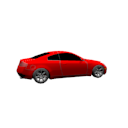} &
\includegraphics[width=1.4cm]{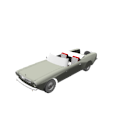} & \includegraphics[width=1.4cm]{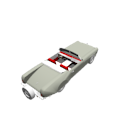} & \includegraphics[width=1.4cm]{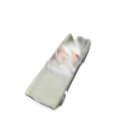} &
\includegraphics[width=1.4cm]{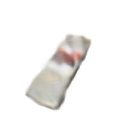} & \includegraphics[width=1.4cm]{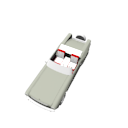} \\
\includegraphics[width=1.4cm]{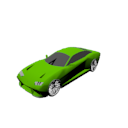} & 
\includegraphics[width=1.4cm]{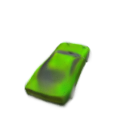} & 
\includegraphics[width=1.4cm]{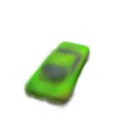} & \includegraphics[width=1.4cm]{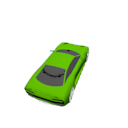} & 
 \includegraphics[width=1.4cm]{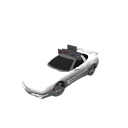} & \includegraphics[width=1.4cm]{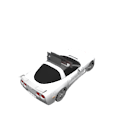}&
 \includegraphics[width=1.4cm]{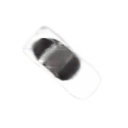} &
 \includegraphics[width=1.4cm]{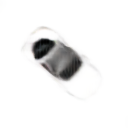} & \includegraphics[width=1.4cm]{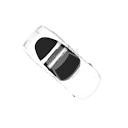}\\
Input & PixelNeRF & Ours   & GT&  \multicolumn{2}{c}{ Inputs} & PixelNeRF & Ours  & GT
\end{tabular}
\end{tabularx}
\vspace{-2mm}
 \caption{Qualitative comparison of novel view synthesis of unseen cars from a single and 2 input views on ShapeNet-V2 \cite{sitzmann2019scene} (More results in \textbf{supp. material}).} 
\label{fig:shapenet}
\vspace{-6mm}
\end{figure}

\subsection{Generalization on Real Data}
\begin{figure}
\vspace{-5mm}
\def\tabularxcolumn#1{m{#1}}
\begin{tabularx}{\linewidth}{@{}cXX@{}}
\setlength{\tabcolsep}{0pt}
\begin{tabular}{cccc|cccc}
\includegraphics[width=1.5cm]{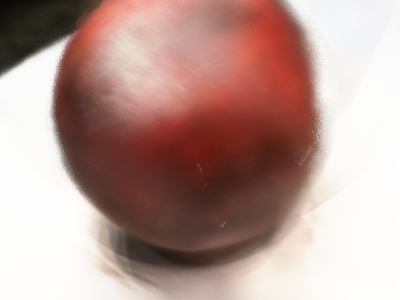} &
\includegraphics[width=1.5cm]{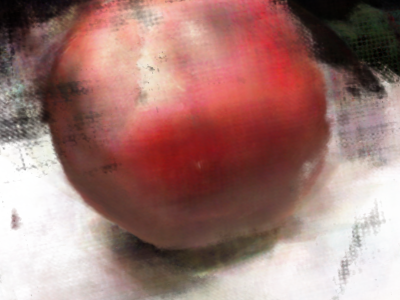} & \includegraphics[width=1.5cm]{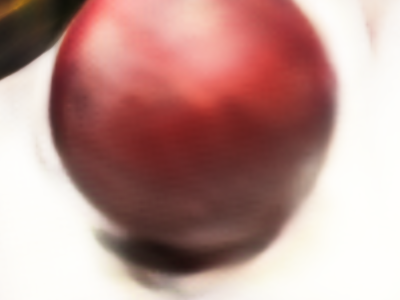} &
\includegraphics[width=1.5cm]{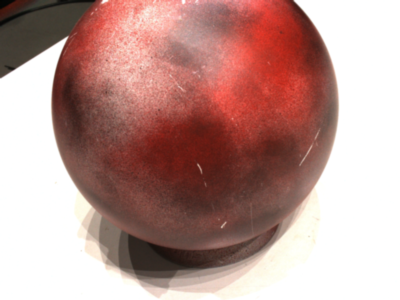} &
\includegraphics[width=1.5cm]{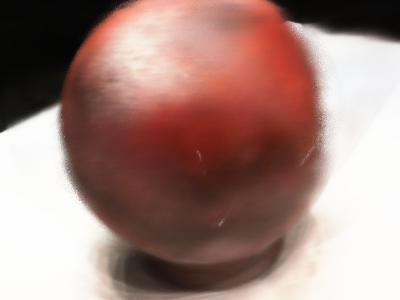} &
\includegraphics[width=1.5cm]{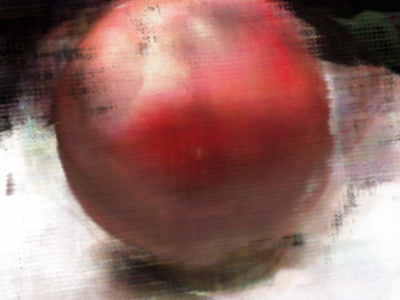} &
\includegraphics[width=1.5cm]{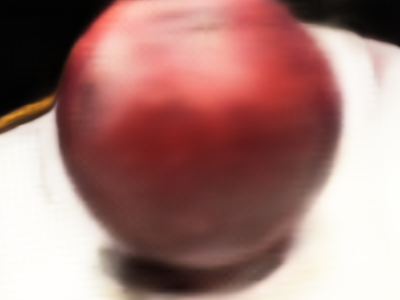} &
\includegraphics[width=1.5cm]{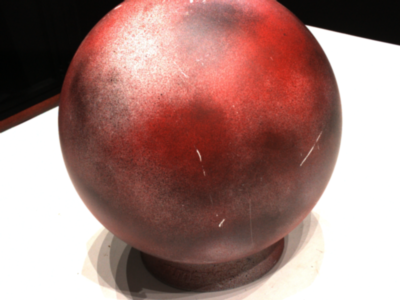} \\
\includegraphics[width=1.5cm]{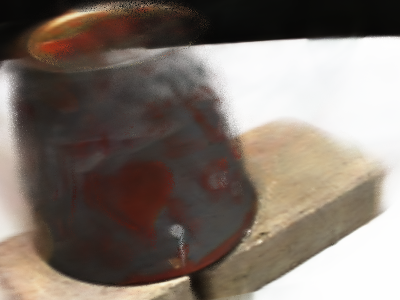} &
\includegraphics[width=1.5cm]{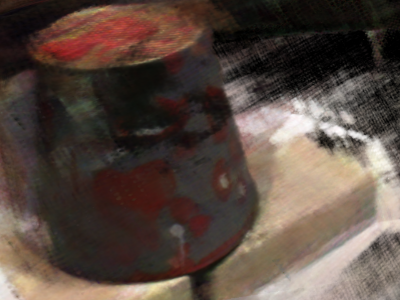} & \includegraphics[width=1.5cm]{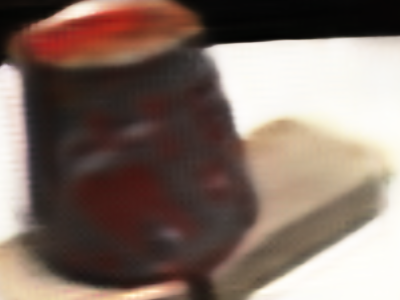} &
\includegraphics[width=1.5cm]{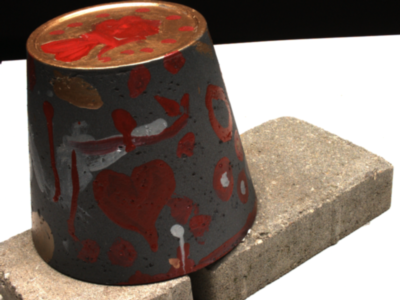} &
\includegraphics[width=1.5cm]{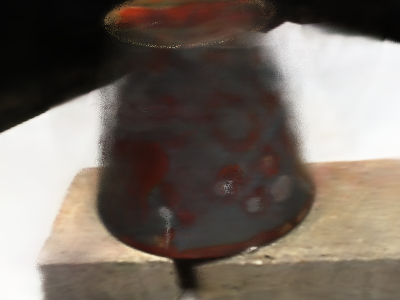} &
\includegraphics[width=1.5cm]{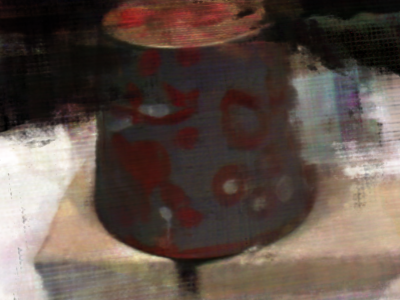} &
\includegraphics[width=1.5cm]{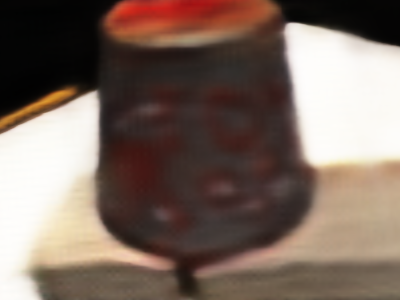} &
\includegraphics[width=1.5cm]{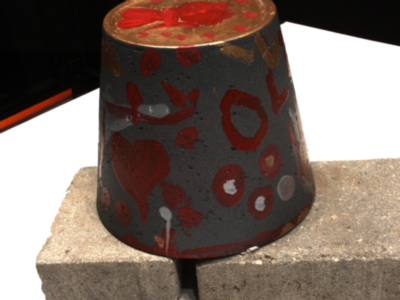} \\
\includegraphics[width=1.5cm]{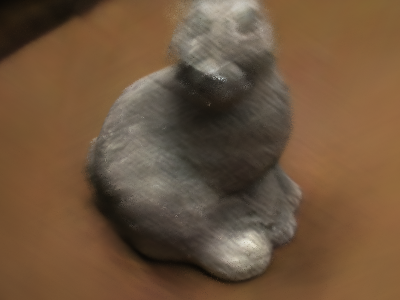} &
\includegraphics[width=1.5cm]{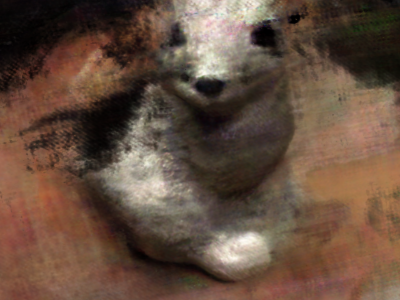} & \includegraphics[width=1.5cm]{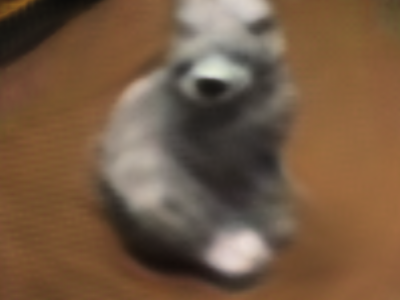} &
\includegraphics[width=1.5cm]{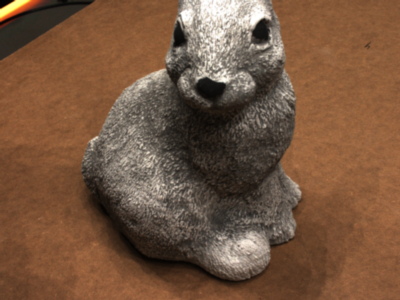} &
\includegraphics[width=1.5cm]{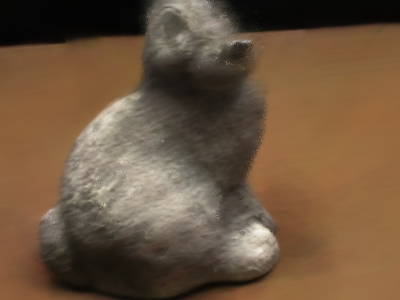} &
\includegraphics[width=1.5cm]{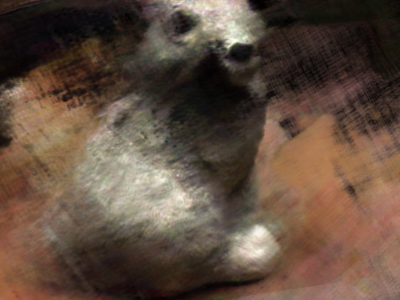} &
\includegraphics[width=1.5cm]{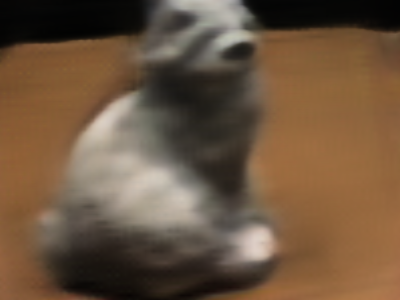} &
\includegraphics[width=1.5cm]{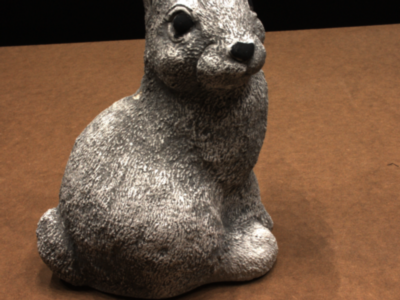} \\
\includegraphics[width=1.5cm]{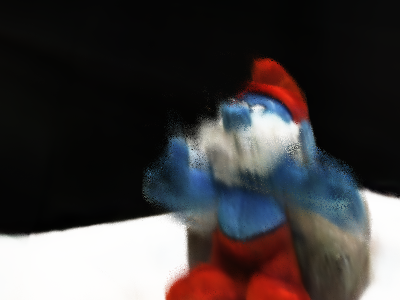} &
\includegraphics[width=1.5cm]{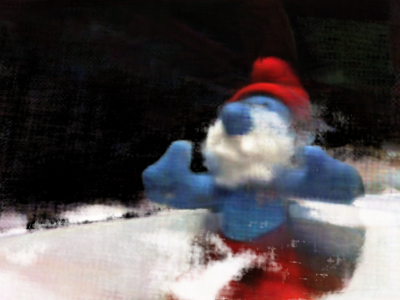} & \includegraphics[width=1.5cm]{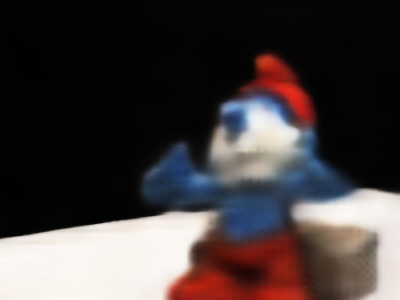} &
\includegraphics[width=1.5cm]{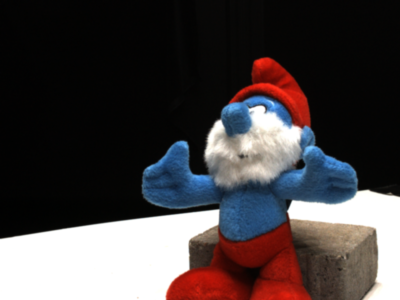} &
\includegraphics[width=1.5cm]{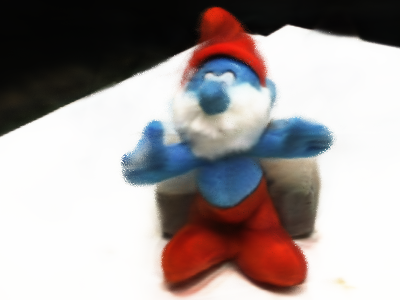} &
\includegraphics[width=1.5cm]{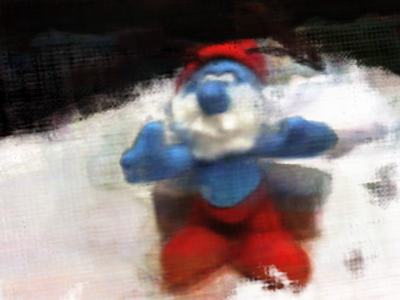} &
\includegraphics[width=1.5cm]{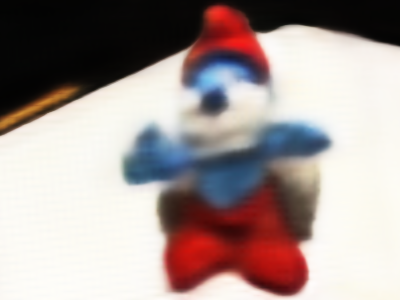} &
\includegraphics[width=1.5cm]{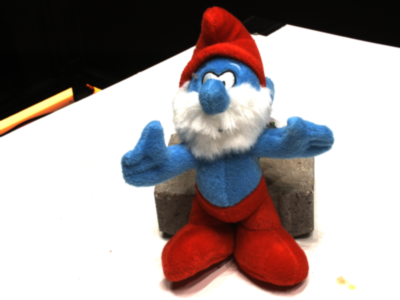} \\
\includegraphics[width=1.5cm]{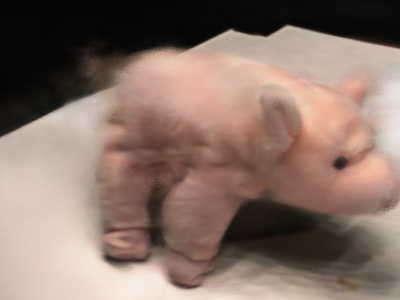} &
\includegraphics[width=1.5cm]{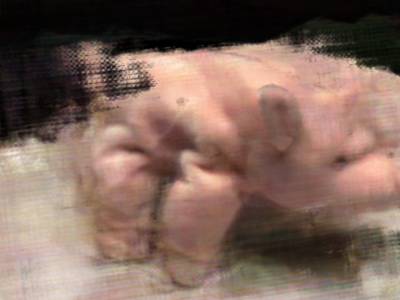} & \includegraphics[width=1.5cm]{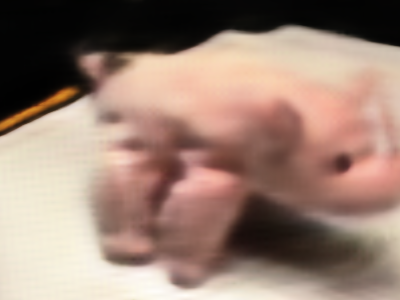} &
\includegraphics[width=1.5cm]{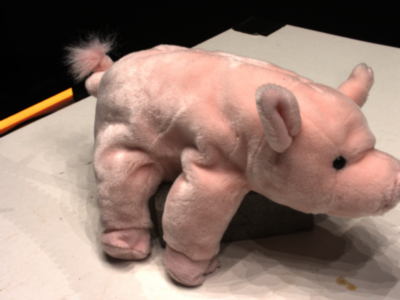} &
\includegraphics[width=1.5cm]{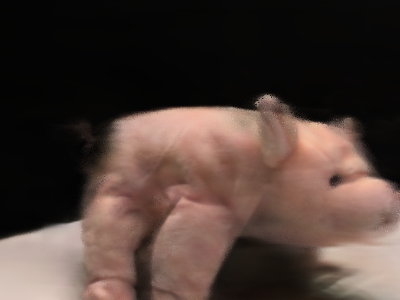} &
\includegraphics[width=1.5cm]{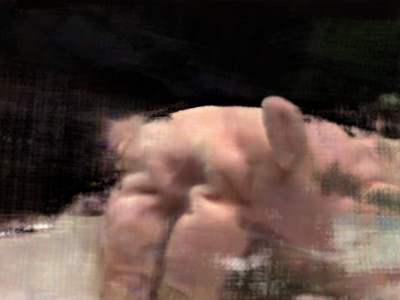} &
\includegraphics[width=1.5cm]{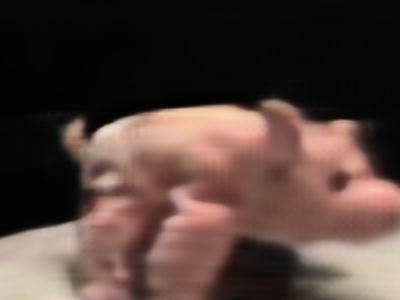} &
\includegraphics[width=1.5cm]{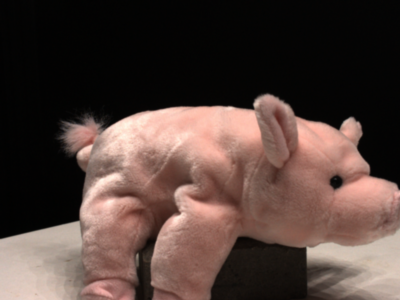} \\
\footnotesize{pixNeRF} & \footnotesize{MVSNeRF} & Ours   & GT & \footnotesize{pixNeRF} & \footnotesize{MVSNeRF} &Ours  & GT
\end{tabular}
\end{tabularx}
 \vspace{-2mm}
 \caption{Qualitative comparison of novel view synthesis of unseen scenes without test time optimization from 6 input views on the DTU dataset \cite{jensen2014large}.} 
\label{fig:dtu} 
 \vspace{-4mm}
\end{figure}

\textbf{Dataset} We demonstrate that our method is capable of reconstructing novel views for real world scenes unseen at training using the DTU dataset \cite{jensen2014large}. Following the PixelNeRF \cite{yu2021pixelnerf} experimental settings, the data is split into 88 training scenes and 16 testing scenes, each scene including 49 images with resolution $300\times 400$. We note that this is a challenging scenario. In fact, the training scenes are limited, and the training and testing scenes do not share any semantic similarities as can be seen in figure \ref{fig:dtu}. The lighting and backgrounds are also inconsistent between the scenes. Hence, this is a few-shot novel view synthesis task that demands considerable scene category generalization as well.\\  

\begin{table}
 \vspace{-6mm}
\begin{center}
\begin{tabular}{l|cc| cc| cc}
\toprule[1.2pt]
Method & \multicolumn{2}{c|} {\bf PSNR$\uparrow$} &  \multicolumn{2}{c|}{\bf SSIM$\uparrow$}& \multicolumn{2}{c}{\bf LPIPS$\downarrow$}\\ &
 3-view & 6-view &3-view  & 6-view & 3-view  & 6-view\\
\hline
SRF\cite{chibane2021stereo} & 15.84  & 17.77 &0.532 & 0.616  & 0.482 & 0.401 \\
PixelNeRF\cite{yu2021pixelnerf} & \cellcolor{shade5}  19.33 & \cellcolor{shade5}  20.43  &\cellcolor{shade5}   0.695 &\cellcolor{shade5}  0.732 &\cellcolor{shade3}  0.387 &\cellcolor{shade3} 0.361 \\
MVSNeRF\cite{chen2021mvsnerf} & \cellcolor{shade1} 16.33 &\cellcolor{shade1} 18.26  &\cellcolor{shade1}  0.602 &\cellcolor{shade3}  0.695  &\cellcolor{shade5}  0.385 &\cellcolor{shade5} 0.321 \\
{\bf Ours} & \cellcolor{shade3}  17.31  & \cellcolor{shade3} 19.20  &\cellcolor{shade3}  0.611 & \cellcolor{shade1} 0.655  & \cellcolor{shade1}  0.433 & \cellcolor{shade1} 0.398  \\
\bottomrule[1.2pt]
\end{tabular}
\caption{Comparison of the average PSNR, SSIM and LPIPS of reconstructed images in the DTU \cite{jensen2014large} dataset \textbf{without test time optimization}. The higher the better for both PSNR and SSIM. The lower the better for LPIPS. Red, orange and blue represent the first, second and third ranking methods respectively.}
\label{tab:dtu}
\end{center}
 \vspace{-12mm}
\end{table}

Table \ref{tab:dtu} shows a quantitative comparison of our method with the recent state-of-the-art in optimization-free few-shot view synthesis. We report the peak signal-to-noise ratio (PSNR), structural similarity (SSIM) and learned perceptual image patch similarity (LPIPS) reconstruction metrics, for the same 3 and 6 view inputs averaged across the same testing scenes. For a fair comparison, we report the performance of PixelNeRF \cite{yu2021pixelnerf} from their paper, and the numbers of methods MVSNeRF \cite{chen2021mvsnerf} and SRF \cite{chibane2021stereo} from RegNeRf \cite{niemeyer2021regnerf}, as the authors in the latter reproduce the performance of these methods in pixelNeRF's DTU setup. Figure \ref{fig:dtu} shows a qualitative comparison between our method and methods pixelNeRF \cite{yu2021pixelnerf} and MVSNerf \cite{chen2021mvsnerf} on synthesized views from testing scenes given the same inputs. We produce the results of PixelNeRF using their publicly available code and DTU model. For MVSNeRF, as their original model was trained on a different DTU setup, we finetune their model on the pixelNeRF DTU setup similarly to RegNeRF \cite{niemeyer2021regnerf}.

Although PixelNeRF and MVSNeRF are based on implicit radiance fields and hence require more evaluations and time for rendering, our implicit light field based method provides competitive performances in comparison. In the single forward prediction setting, our method is overall second to PixelNeRF in PSNR, while providing a 100 times faster inference  (see table \ref{tab:time}).  As illustrated in the visual comparison in figure \ref{fig:dtu}, we manage to reproduce the shape and appearance of the scene to a good extent and also recover from some of the competitions' failures. Our method appears to be better in fact at preserving the coarser structure of the scene. Specifically, some elements of the ground truth that we manage to reproduce are not recovered by the competition, such as the pig tail in the last row, the background table's yellow lines and the rabbit's eyes. Although we found MVSNeRF encounters multiple failures compared to pixelNeRF, it is apparent that NeRF base methods (pixelNeRF and MVSNeRF) are able to produce some relatively higher frequency details, albeit at a considerable higher rendering cost. 

\begin{table}
 \vspace{-5mm}
\begin{center}
\begin{tabular}{l|cc| cc| cc}
\toprule[1.2pt]
Method & \multicolumn{2}{c|} {\bf PSNR$\uparrow$} &  \multicolumn{2}{c|}{\bf SSIM$\uparrow$}& \multicolumn{2}{c}{\bf LPIPS$\downarrow$}\\ &
3-view & 6-view &3-view  & 6-view & 3-view  & 6-view\\
\hline
mip-NeRF\cite{barron2021mip}  & 7.64 & 14.33 & 0.227 &0.568  &0.655 &0.394\\
DietNeRF \cite{jain2021putting} & 10.01 & 18.70 & 0.354 & 0.668 & 0.574 &\cellcolor{shade1}  0.336 \\
RegNeRF \cite{niemeyer2021regnerf} & 15.33 &\cellcolor{shade1}  19.10  &\cellcolor{shade3}  0.621 &\cellcolor{shade5}  0.757&\cellcolor{shade5}  0.341 &\cellcolor{shade5}  0.233 \\
\hline
SRF \cite{chibane2021stereo} & 16.06  & 18.69 & 0.550 & 0.657  & 0.431 & 0.353 \\
PixelNeRF\cite{yu2021pixelnerf} &\cellcolor{shade3}  17.38 &\cellcolor{shade5}  21.52  & 0.548 &\cellcolor{shade1}  0.670  & 0.456 &0.351 \\
MVSNeRF\cite{chen2021mvsnerf} & \cellcolor{shade1}  16.26 & 18.22  &\cellcolor{shade1}  0.601 & 0.694 &\cellcolor{shade3}  0.384 &\cellcolor{shade3}  0.319 \\
{\bf Ours} &\cellcolor{shade5}  17.72  &\cellcolor{shade3} 19.56 &\cellcolor{shade5}  0.626 &\cellcolor{shade3}  0.671  &\cellcolor{shade1}  0.412 & 0.375  \\
\bottomrule[1.2pt]
\end{tabular}
\caption{Comparison of the average PSNR, SSIM and LPIPS of reconstructed images in the DTU \cite{jensen2014large} dataset \textbf{with test time optimization}. The higher the better for both PSNR and SSIM. The lower the better for LPIPS. Red, orange and blue represent the first, second and third ranking methods respectively.}
\label{tab:dtu-ft}
\end{center}
 \vspace{-10mm}
\end{table}

\begin{figure}
\vspace{-5mm}
\flushleft 
\def\tabularxcolumn#1{m{#1}}
\begin{tabularx}{\linewidth}{@{}cXX@{}}
\setlength{\tabcolsep}{0pt}
\begin{tabular}{cccc|cccc}
\includegraphics[width=1.5cm]{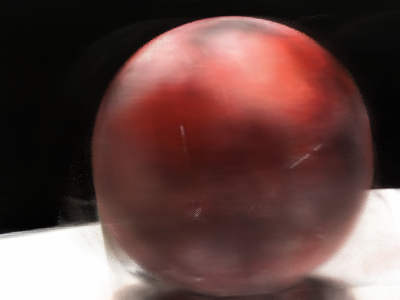} &
\includegraphics[width=1.5cm]{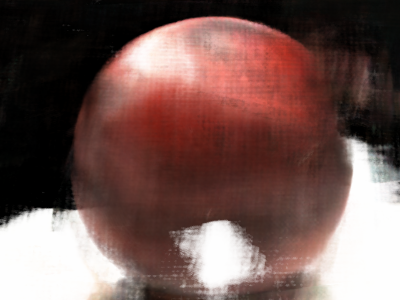} & 
\includegraphics[width=1.5cm]{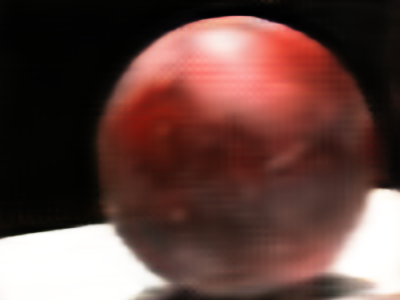} &
\includegraphics[width=1.5cm]{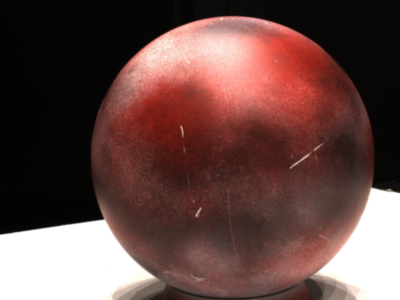} &
\includegraphics[width=1.5cm]{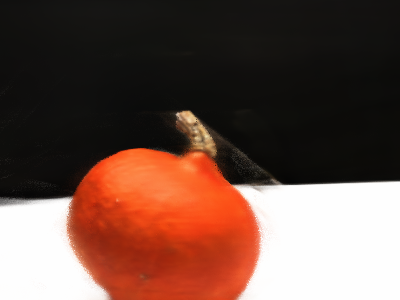} &
\includegraphics[width=1.5cm]{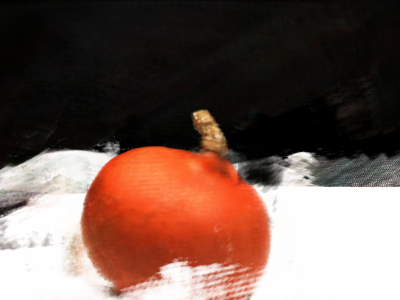} &
\includegraphics[width=1.5cm]{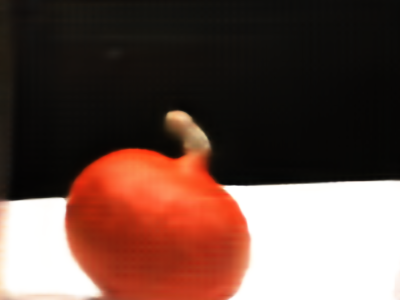} &
\includegraphics[width=1.5cm]{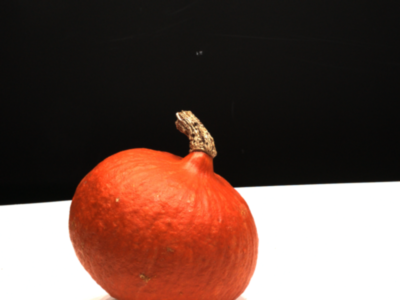}\\
\includegraphics[width=1.5cm]{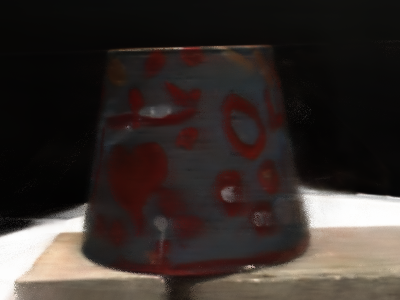} &
\includegraphics[width=1.5cm]{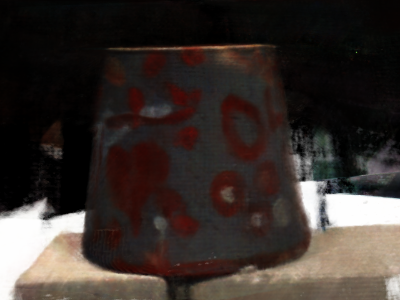} & \includegraphics[width=1.5cm]{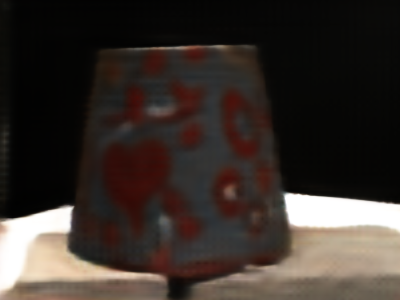} &
\includegraphics[width=1.5cm]{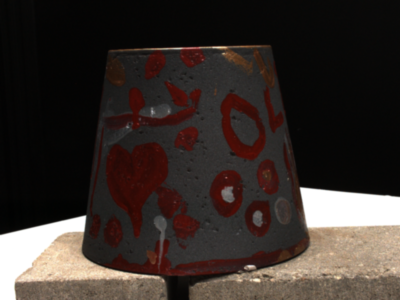} &
\includegraphics[width=1.5cm]{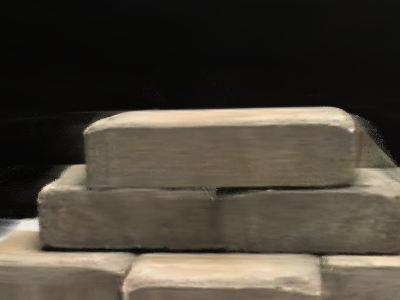} &
\includegraphics[width=1.5cm]{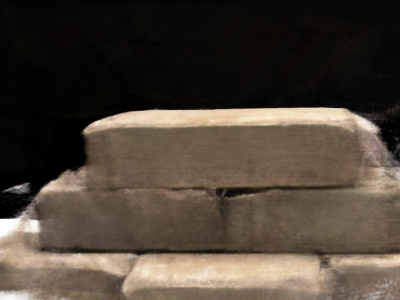} &
\includegraphics[width=1.5cm]{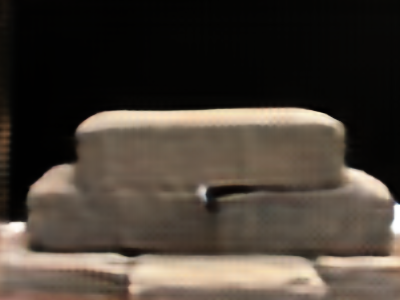} &
\includegraphics[width=1.5cm]{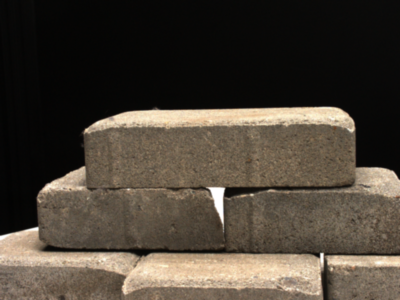}\\
\includegraphics[width=1.5cm]{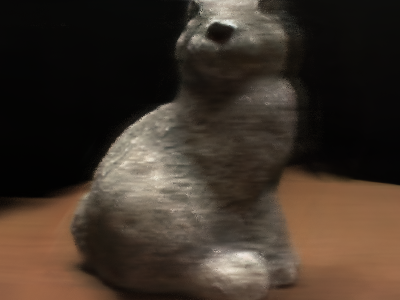} &
\includegraphics[width=1.5cm]{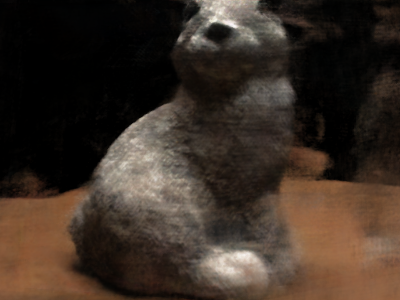} & \includegraphics[width=1.5cm]{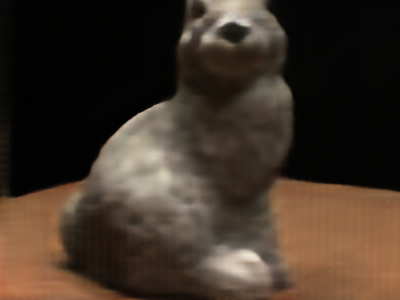} &
\includegraphics[width=1.5cm]{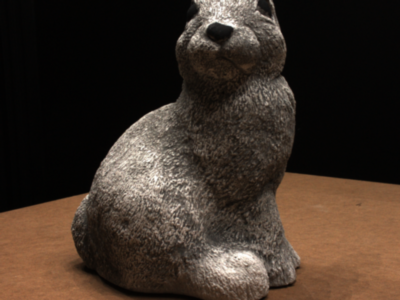} &
\includegraphics[width=1.5cm]{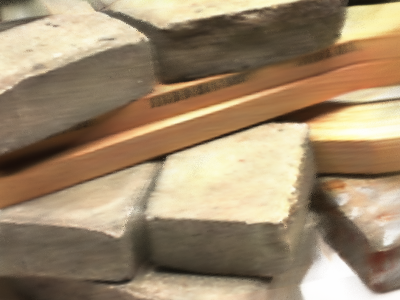} &
\includegraphics[width=1.5cm]{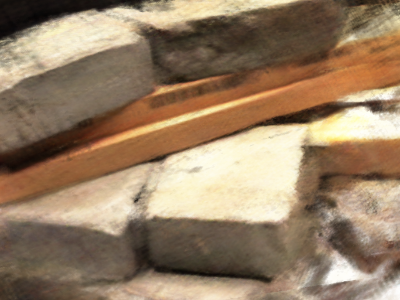} &
\includegraphics[width=1.5cm]{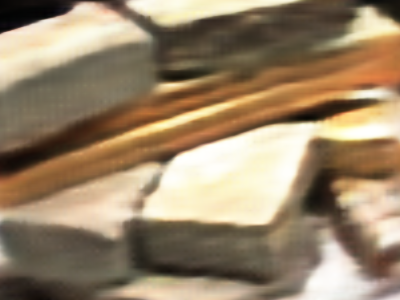} &
\includegraphics[width=1.5cm]{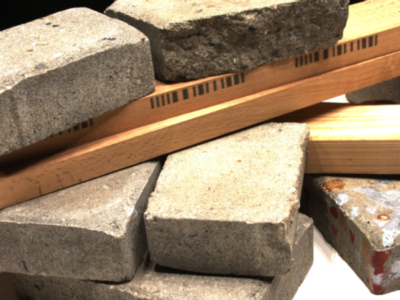} \\
\includegraphics[width=1.5cm]{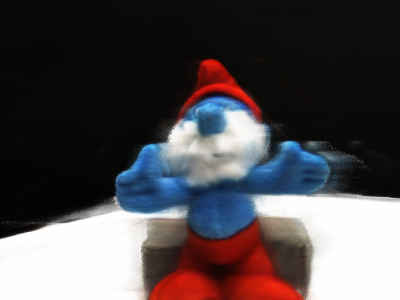} &
\includegraphics[width=1.5cm]{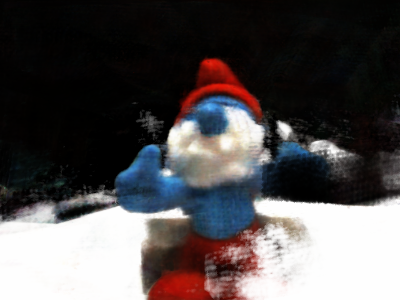} &
\includegraphics[width=1.5cm]{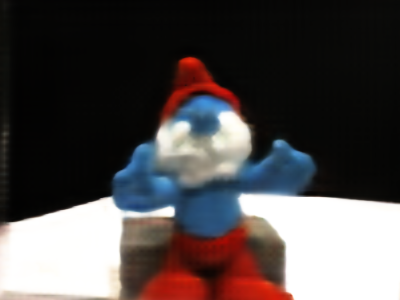} &
\includegraphics[width=1.5cm]{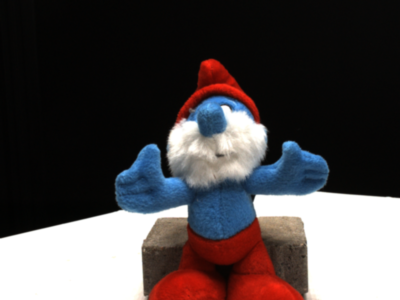} &
\includegraphics[width=1.5cm]{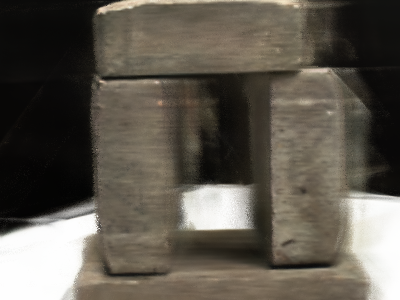} &
\includegraphics[width=1.5cm]{fig/dtu/finetune6views/scan40/43_pl.png} & \includegraphics[width=1.5cm]{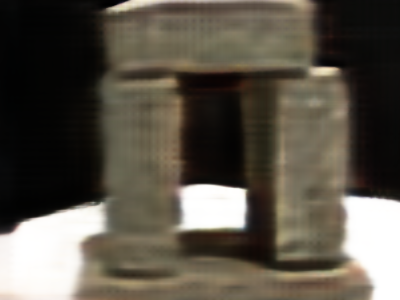} &
\includegraphics[width=1.5cm]{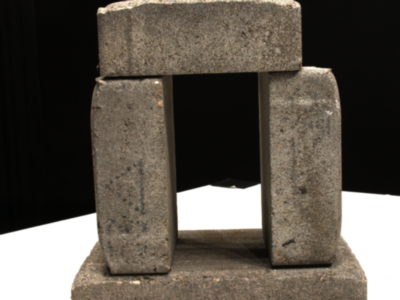} \\
\includegraphics[width=1.5cm]{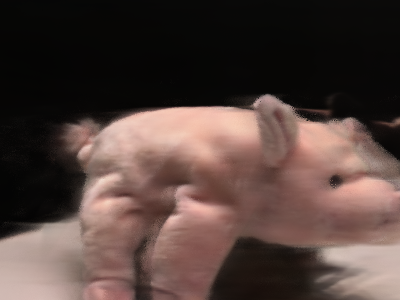} &
\includegraphics[width=1.5cm]{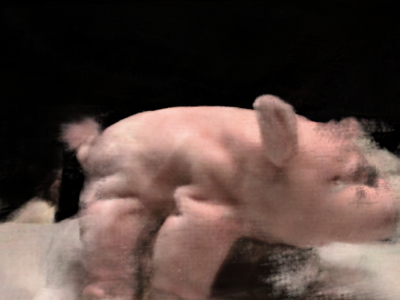} & \includegraphics[width=1.5cm]{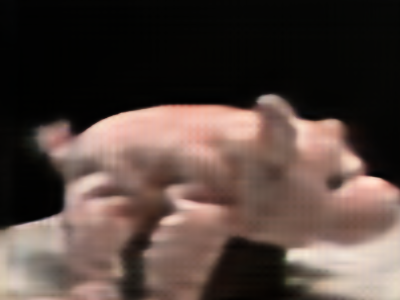} &
\includegraphics[width=1.5cm]{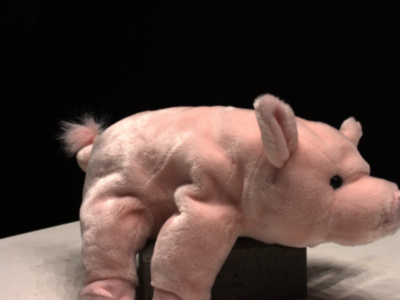} &
\includegraphics[width=1.5cm]{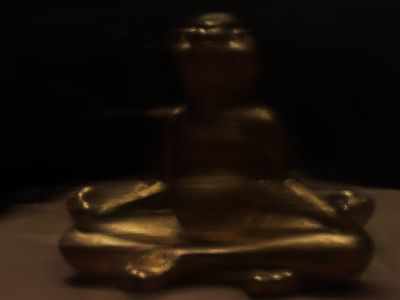} &
\includegraphics[width=1.5cm]{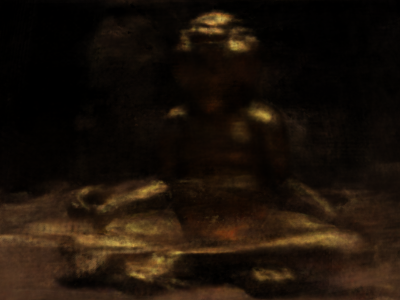} &
\includegraphics[width=1.5cm]{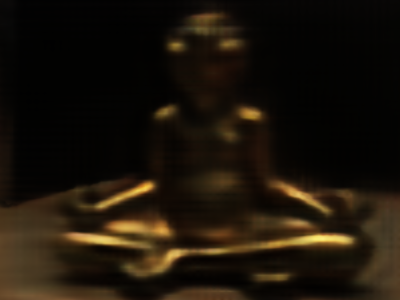} &
\includegraphics[width=1.5cm]{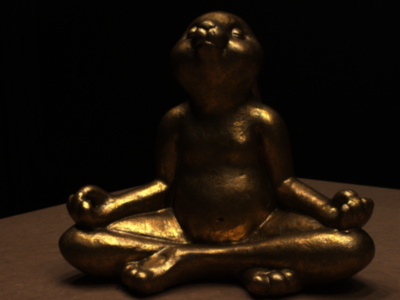} \\
 \footnotesize{pixNeRF} & \footnotesize{MVSNeRF} & Ours   & GT & \footnotesize{pixNeRF} & \footnotesize{MVSNeRF} &Ours  & GT
\end{tabular}
&
\end{tabularx}
\vspace{-2mm}
 \caption{Qualitative comparison of novel view synthesis with test time optimization using 6 input views on the DTU dataset \cite{jensen2014large}.}
 \label{fig:dtu-ft}
 \vspace{-4mm}
\end{figure}

Table \ref{tab:dtu-ft} shows a quantitative comparison of our method with the recent few-shot novel view synthesis state-of-the-art with test time optimization. We report PSNR, SSIM and LPIPS for the same 3 and 6 input views averaged over the same testing objects. All the method's numbers on the PixelNeRF DTU setup are reported as reproduced in RegNeRF \cite{niemeyer2021regnerf}. Methods mip-NeRF \cite{barron2021mip}, DietNeRF \cite{jain2021putting} and RegNeRF \cite{niemeyer2021regnerf} are optimized per scene only, while PixelNeRF \cite{yu2021pixelnerf}, MVSNeRF \cite{chen2021mvsnerf} and ours are trained on the DTU training set then finetuned per scene. Following the experimental setting in  RegNeRF, only the input views were used for fine-tuning. Similarly to the finetuning of MVSNeRf and PixelNeRF, we reduce the learning rate from $10^{-4}$ to $10^{-5}$ and constrain the finetuning within 10k iterations for better performance.  
Figure \ref{fig:dtu-ft} shows a qualitative comparison to MVSNeRf and PixelNeRF given 6 input views after finetuning.

In the 3 input view case, we outperform all methods in the PSNR and SSIM metrics. We obtain overall comparable performances with generalizable (PixelNeRF and MVSNeRF) and single scene optimization (RegNeRF) NeRFs. Figure \ref{fig:dtu-ft} shows that we can achieve relatively comparable results to the encoder endowed NeRF approaches after optimization. We recall again that competition methods here require renderings that are orders of magnitude slower than ours. 

\subsection{Ablation}

We propose here an ablative analysis for our method on the DTU \cite{jensen2014large} and shapeNet-V2 \cite{sitzmann2019scene} datasets. Specifically, we disable the light field function (ours w/o light field), and we render the final image directly from the target view aligned convolutional feature volume. We also reproduce our method without using the ray coordinates (ours w/o ray coordinates). 

\begin{figure}
 \vspace{-5mm}
\flushleft 
\def\tabularxcolumn#1{m{#1}}
\begin{tabularx}{\linewidth}{@{}cXX@{}}
\setlength{\tabcolsep}{0pt}
\begin{tabular}{cc cc cc cc cc}
\includegraphics[width=1.2cm]{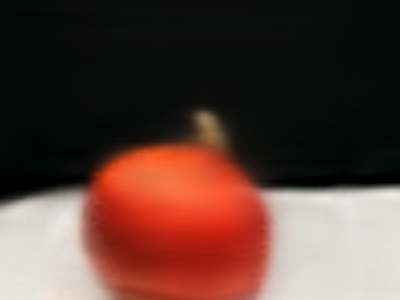} &
\includegraphics[width=1.2cm]{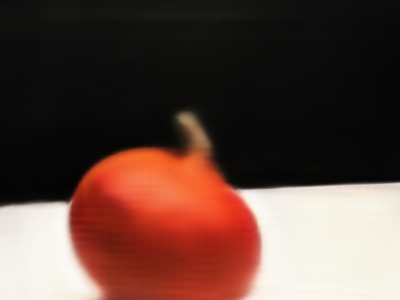} &
\includegraphics[width=1.2cm]{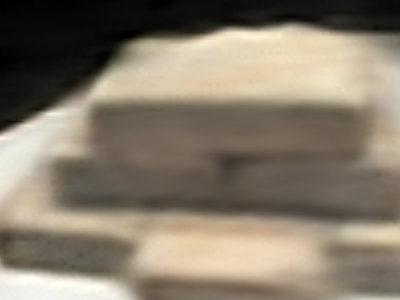} &
\includegraphics[width=1.2cm]{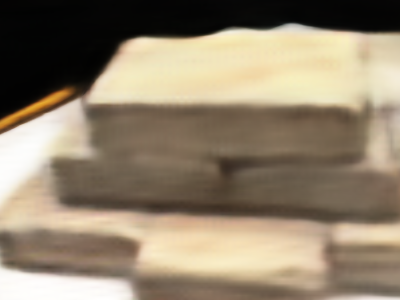} &
\includegraphics[width=1.2cm]{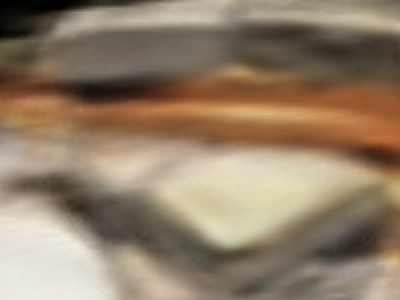} &
\includegraphics[width=1.2cm]{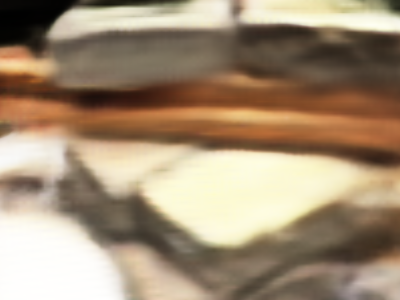} &
\includegraphics[width=1.2cm]{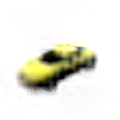} &
\includegraphics[width=1.2cm]{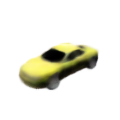} &
\includegraphics[width=1.2cm]{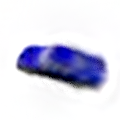} &
\includegraphics[width=1.2cm]{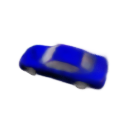} 
\\
\includegraphics[width=1.2cm]{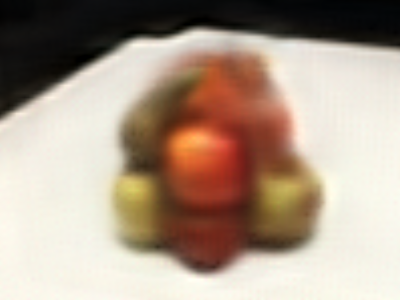} &
\includegraphics[width=1.2cm]{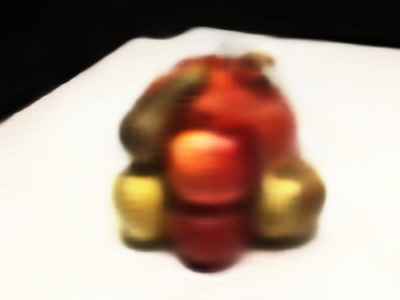} &
\includegraphics[width=1.2cm]{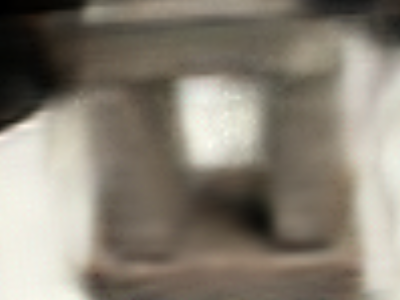} &
\includegraphics[width=1.2cm]{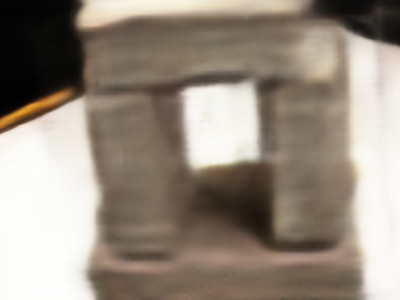} &
\includegraphics[width=1.2cm]{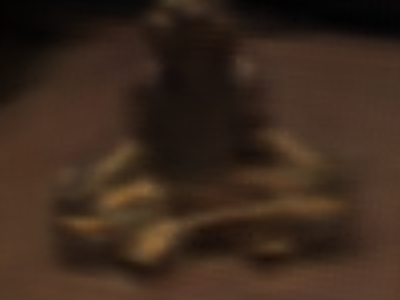}&
\includegraphics[width=1.2cm]{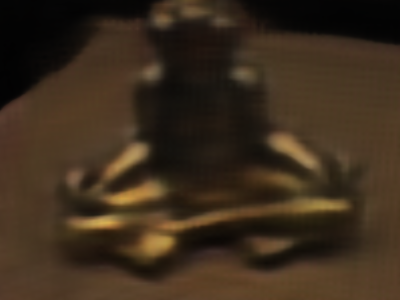}&
\includegraphics[width=1.2cm]{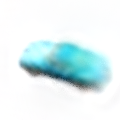} &
\includegraphics[width=1.2cm]{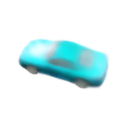} &
\includegraphics[width=1.2cm]{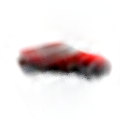} &
\includegraphics[width=1.2cm]{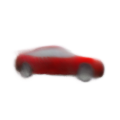} 
\\
w/o lf &\textbf{Ours} &w/o lf &\textbf{Ours}  &w/o lf  &\textbf{Ours}&w/o lf  &\textbf{Ours}&w/o lf  &\textbf{Ours}
\end{tabular}
\end{tabularx}
 \vspace{-2mm}
 \caption{Qualitative ablation of our method on unseen DTU \cite{jensen2014large} scenes (6 input views) and unseen ShapeNet-V2\cite{sitzmann2019scene} cars (1 input view).}
 \label{fig:abl} 
\end{figure}

Table \ref{tab:abl} shows numerical comparisons for 3 and 6 input views  on DTU, and figure \ref{fig:abl}
shows qualitative comparisons for 6 input views on DTU, and 1 input view on ShapeNet-V2. Our final method improves on its 3D aware convolutional baseline both numerically and visually. In particular, we note how the light field network with ray encoding reduces ghosting artifacts and provides finer details.


\begin{table}
\begin{center}
\begin{tabular}{l|cc| cc| cc}
\toprule[1.2pt]
Method  & \multicolumn{2}{c|} {\bf PSNR$\uparrow$} &  \multicolumn{2}{c|}{\bf SSIM$\uparrow$}& \multicolumn{2}{c}{\bf LPIPS$\downarrow$}\\ 
 & 3-view & 6-view &3-view  & 6-view & 3-view  & 6-view \\
\hline
 Ours w/o light field  & 16.61  & 17.30    &0.575 & 0.603 & 0.462 & 0.402 \\
 Ours w/o ray coordinates  & 16.77  & 17.42    &0.604 & 0.636 & 0.455 & 0.403 \\
\textbf{Ours}  & \textbf{17.31}  & \textbf{19.20}  & \textbf{0.611} & \textbf{0.655}  & \textbf{0.433}  & \textbf{0.398} \\
\bottomrule[1.2pt]
\end{tabular}
\caption{Quantitative ablation of our method on the DTU dataset \cite{jensen2014large}. The higher the better for both PSNR and SSIM. The lower the better for LPIPS.}
\label{tab:abl}
\end{center}
\end{table}

\section{Limitations}

Whilst our method offers an efficient rendering (100 times faster than PixelNeRF), it still has difficulties in reproducing the highest level of details in real large images as the NeRF based methods. We believe this is due to the reduced resolution of our feature volume and our coarse feature rendering, which conversely contribute towards reducing the memory footprint. As future work, we will attempt to overcome this limitation while maintaining memory efficiency. Furthermore, the explicit feature volume limits the applicability of our method on datasets with larger images such as LLFF \cite{mildenhall2019local}. We will thus consider intrinsic 3D feature representations next.

\section{Conclusion}
We proposed a method for generating novel views from few input calibrated images with a single forward pass prediction deep neural network. We learn an implicit neural light field function that models ray colors directly. In comparison to \cite{sitzmann2021light}, we proposed a more efficient local ray conditioning, and an optimization free inference. Our method combines the advantages of 3D aware convolutional approaches and implicit representations, and requires only image data in training. We demonstrated our method successfully on synthetic and real benchmarks for few-shot novel view synthesis. 
Our method outperforms the convolutional baselines (see table \ref{tab:shapenet}) and provides competitive performances compared to locally conditioned radiance fields (\eg PixelNeRF \cite{yu2021pixelnerf}), while being a $100$ times faster to render.      



\bibliographystyle{splncs}
\bibliography{egbib}



\section*{Supplementary Material}

\subsection*{Network architecture}
Table \ref{tab:e} describes the detailed architecture of our convolutional network $E$ introduced in Section 3.1 of the main submission. Table \ref{tab:f} shows the detailed architecture of our network $f$ introduced in Section 3.5 in the main submission.   

\subsection*{DTU \cite{jensen2014large} evaluation protocol}
We follow the evaluation setup in PixelNeRF \cite{yu2021pixelnerf,niemeyer2021regnerf} and use the following scans as the test set: 8, 21, 30, 31,
34, 38, 40, 41, 45, 55, 63, 82, 103, 110, 114. The following images are used as input: 25, 22, 28, 40, 44, 48, 0, 8, 13. For the 3 and 6 input scenarios, we use the first 3/6 images from the list. All remaining images are used for evaluation except wrong exposure images 3, 4, 5, 6, 7, 16, 17, 18, 19, 20, 21,
36, 37, 38, 39. The input image resolution is 300 $\times$ 400.

\subsection*{Additional qualitative results}
\noindent\textbf{Results on ShapeNet-V2 \cite{sitzmann2019scene}}\quad Figures \ref{fig:chair2}, \ref{fig:chair1}, \ref{fig:car2}, \ref{fig:car1} show additional novel view synthesis qualitative results of our method using a single and 2 input views.\\
\noindent\textbf{Comparison to LFN \cite{sitzmann2021light}}\quad
Figure \ref{fig:chairs-lfn} shows additional comparisons to Sitzmann et al. \cite{sitzmann2021light} on the chair class of ShapeNet-V2 \cite{sitzmann2019scene}.

\begin{table}
\flushleft
\begin{center}
 \setlength{\tabcolsep}{8pt}
 \renewcommand{\arraystretch}{1}
\begin{tabular}{c c c c}
\toprule[1.2pt]
\hline
{\bf Input Shape} & {\bf Output shape} &{\bf Operation}  &\\
\hline
$(3,H,W)$&  $(52,H,W)$ & 1$\times$1 Conv & Image\\
\hline
$(12,H,W)$& $(12,H,W)$ & 1$\times$1 Conv & Relative Pose\\
\hline
$(52+12,H,W)$&  $(64,H,W)$ &  1$\times$1 Conv & \multirow{10}{*}{2D Conv}\\
$(64,H,W)$&  $(64,H,W)$ & 2$\times$ ResBlock &~\\
$(64,H,W)$&   $(128,H/2,W/2)$ & 4$\times$4 Conv,Stride2 &~\\
$(128,H/2,W/2)$&  $(128,H/2,W/2)$ & 1$\times$ ResBlock &~\\
$(128,H/2,W/2)$&  $(128,H/4,W/4)$ & 4$\times$4 Conv,Stride2 &~\\
$(128,H/4,W/4)$ &  $(128,H/4,W/4)$ & 1$\times$ ResBlock &~\\
$(128,H/4,W/4)$& $(256,H/8,W/8)$ & 4$\times$4 Conv,Stride2 &~\\
$(256,H/8,W/8)$&   $(256,H/8,W/8)$ & 1$\times$ ResBlock &~\\
$(256,H/8,W/8)$ &  $(128,H/4,W/4)$  & 4$\times$4 Conv.T,Stride2 &~\\
$(128,H/4,W/4)$  &  $(128,H/4,W/4)$  & 2$\times$ ResBlock &~\\
\hline
$(128,H/4,W/4)$ &  $(256,H/4,W/4)$  & 1$\times$1 Conv & \multirow{4}{*}{2D to 3D}\\
 $(256,H/4,W/4)$ &    $(512,H/4,W/4)$  & 1$\times$1 Conv  &~\\
 $(512,H/4,W/4)$ &   $(2048,H/4,W/4)$   & 1$\times$1 Conv  &~\\
 $(2048,H/4,W/4)$ &   $(32,64,H/4,W/4)$   & Reshape &~\\
\hline
 $(32,64,H/4,W/4)$&    $(32,64,H/4,W/4)$ & 1$\times$1 Conv & \multirow{6}{*}{3D Conv}\\
 $(32,64,H/4,W/4)$&    $(32,64,H/4,W/4)$  & 2$\times$ ResBlock &~\\
 $(32,64,H/4,W/4)$&    $(64,32,H/8,W/8)$  & 4$\times$4 Conv,Stride2 &~\\
$(64,32,H/8,W/8)$&   $(64,32,H/8,W/8)$  & 2$\times$ ResBlock &~\\
$(64,32,H/8,W/8)$&   $(32,64,H/4,W/4)$  & 4$\times$4 Conv.T,Stride2 &~\\
 $(32,64,H/4,W/4)$&  $(32,64,H/4,W/4)$  & 2$\times$ ResBlock &~\\
 \hline
  $(31,64,H/4,W/4)$&  $(31,H/4,W/4)$  & - &Rendering\\
 \hline
 $(30,H/4,W/4)$&  $(30,H/2,W/2)$  &  1$\times$1 Conv,Upsampling &\multirow{2}{*}{Upsampling}\\
$(30,H/2,W/2)$&  $(30,H,W)$ & 1$\times$1 Conv,Upsampling &\\
\bottomrule[1.2pt]
\end{tabular}
\caption{The architecture of network $E$ in Section 3.1 of the main submission.}
\label{tab:e}
\end{center}
\end{table}

\begin{table}
\begin{center}
\setlength{\tabcolsep}{12pt}
 \renewcommand{\arraystretch}{1.2}
\begin{tabular}{c c c}
\toprule[1.2pt]
\hline
{\bf Layer} & {\bf Channels}   &{\bf Inputs}\\
\hline
$LR_0$&  30/256 &$\mathcal{F}_{u,v}$\\ 
$PE$&  6/78 &$r_{u,v}$\\
$LR_1$&  78/256  &$PE$\\
$LR_2$&  256/256 &$LR_1 \odot LR_0 $\\
$LR_3$&  256/256 &$LR_2 \odot LR_0 $\\
$LR_4$&  256/256 &$LR_3 \odot LR_0 $\\
$LR_5$&  256/256 &$LR_4 \odot LR_0 $\\
$LR_6$&  256/256 &$LR_5 \odot LR_0 $\\
$c_{u,v}$&  256/3  &$LR_6$\\
\bottomrule[1.2pt]
\end{tabular}
\caption{The MLP in Section 3.5 of the main submission. 
$r_{u,v}$ is the Plücker coordinate of the ray. $\mathcal{F}_{u,v}$ is the feature of the ray. $c_{u,v}$ is the target pixel color.  
$PE$ is the positional encoding in \cite{mildenhall2020nerf}. $\odot$ is the element-wise product. All layers are linear with Relu activation, except the last layer which uses a Sigmoid.}
\label{tab:f}
\end{center}
\end{table}

\begin{figure}
\def\tabularxcolumn#1{m{#1}}
\begin{tabularx}{\linewidth}{@{}cXX@{}}
\setlength{\tabcolsep}{0pt}
\begin{tabular}{c|c cc cc cc cc}
\multicolumn{1} {c|}{\bf Inputs} &  \multicolumn{9}{c}{\bf Rendered novel views}\\
\includegraphics[width=1.2cm]{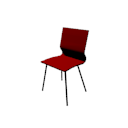} &
\includegraphics[width=1.2cm]{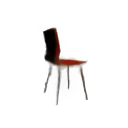} &
\includegraphics[width=1.2cm]{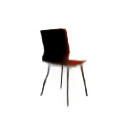} &
\includegraphics[width=1.2cm]{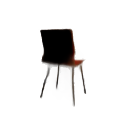} &
\includegraphics[width=1.2cm]{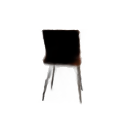} &
\includegraphics[width=1.2cm]{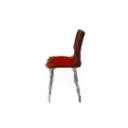} &
\includegraphics[width=1.2cm]{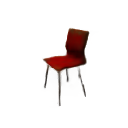} &
\includegraphics[width=1.2cm]{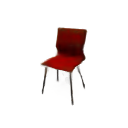} &
\includegraphics[width=1.2cm]{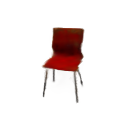}&
\includegraphics[width=1.2cm]{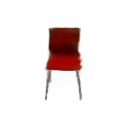}
\\
\includegraphics[width=1.2cm]{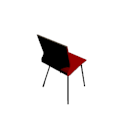} &
\includegraphics[width=1.2cm]{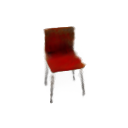} &
\includegraphics[width=1.2cm]{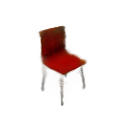} &
\includegraphics[width=1.2cm]{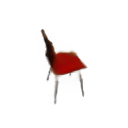} &
\includegraphics[width=1.2cm]{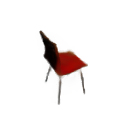} &
\includegraphics[width=1.2cm]{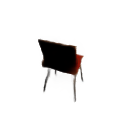} &
\includegraphics[width=1.2cm]{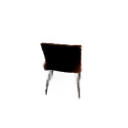} &
\includegraphics[width=1.2cm]{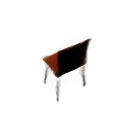} &
\includegraphics[width=1.2cm]{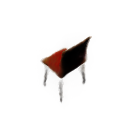}&
\includegraphics[width=1.2cm]{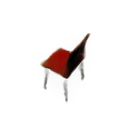} \\
\hline

\includegraphics[width=1.2cm]{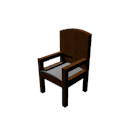} &
\includegraphics[width=1.2cm]{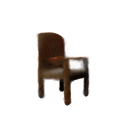} &
\includegraphics[width=1.2cm]{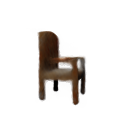} &
\includegraphics[width=1.2cm]{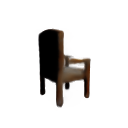} &
\includegraphics[width=1.2cm]{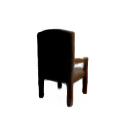} &
\includegraphics[width=1.2cm]{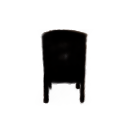} &
\includegraphics[width=1.2cm]{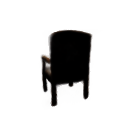} &
\includegraphics[width=1.2cm]{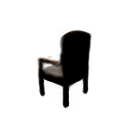} &
\includegraphics[width=1.2cm]{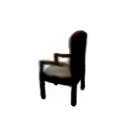}&
\includegraphics[width=1.2cm]{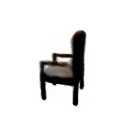}
\\
\includegraphics[width=1.2cm]{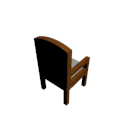} &
\includegraphics[width=1.2cm]{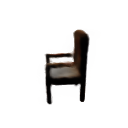} &
\includegraphics[width=1.2cm]{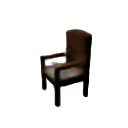} &
\includegraphics[width=1.2cm]{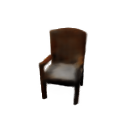} &
\includegraphics[width=1.2cm]{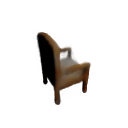} &
\includegraphics[width=1.2cm]{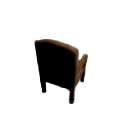} &
\includegraphics[width=1.2cm]{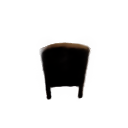} &
\includegraphics[width=1.2cm]{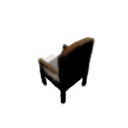} &
\includegraphics[width=1.2cm]{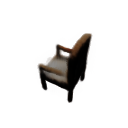}&
\includegraphics[width=1.2cm]{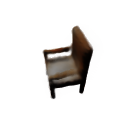} \\
\hline

\includegraphics[width=1.2cm]{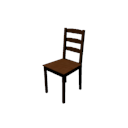} &
\includegraphics[width=1.2cm]{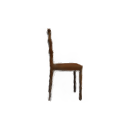} &
\includegraphics[width=1.2cm]{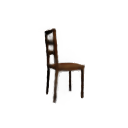} &
\includegraphics[width=1.2cm]{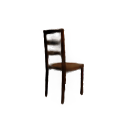} &
\includegraphics[width=1.2cm]{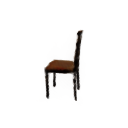} &
\includegraphics[width=1.2cm]{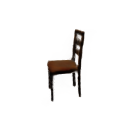} &
\includegraphics[width=1.2cm]{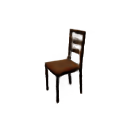} &
\includegraphics[width=1.2cm]{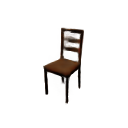} &
\includegraphics[width=1.2cm]{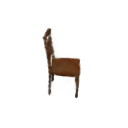}&
\includegraphics[width=1.2cm]{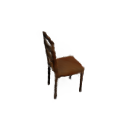}
\\
\includegraphics[width=1.2cm]{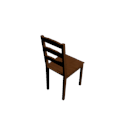} &
\includegraphics[width=1.2cm]{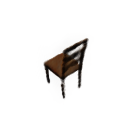} &
\includegraphics[width=1.2cm]{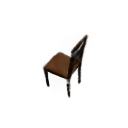} &
\includegraphics[width=1.2cm]{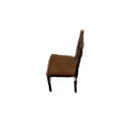} &
\includegraphics[width=1.2cm]{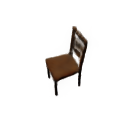} &
\includegraphics[width=1.2cm]{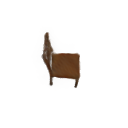} &
\includegraphics[width=1.2cm]{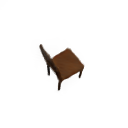} &
\includegraphics[width=1.2cm]{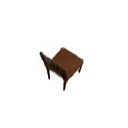} &
\includegraphics[width=1.2cm]{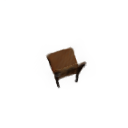}&
\includegraphics[width=1.2cm]{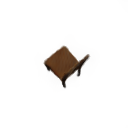} \\
\hline

\includegraphics[width=1.2cm]{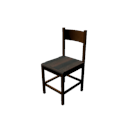} &
\includegraphics[width=1.2cm]{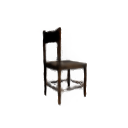} &
\includegraphics[width=1.2cm]{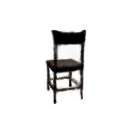} &
\includegraphics[width=1.2cm]{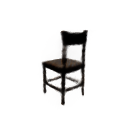} &
\includegraphics[width=1.2cm]{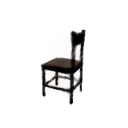} &
\includegraphics[width=1.2cm]{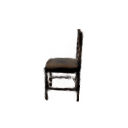} &
\includegraphics[width=1.2cm]{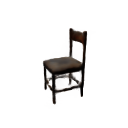} &
\includegraphics[width=1.2cm]{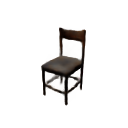} &
\includegraphics[width=1.2cm]{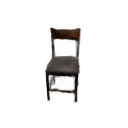}&
\includegraphics[width=1.2cm]{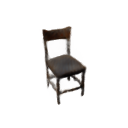}
\\
\includegraphics[width=1.2cm]{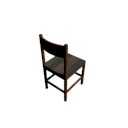} &
\includegraphics[width=1.2cm]{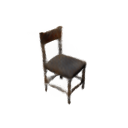} &
\includegraphics[width=1.2cm]{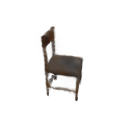} &
\includegraphics[width=1.2cm]{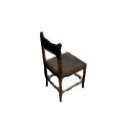} &
\includegraphics[width=1.2cm]{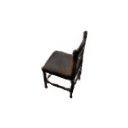} &
\includegraphics[width=1.2cm]{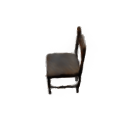} &
\includegraphics[width=1.2cm]{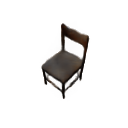} &
\includegraphics[width=1.2cm]{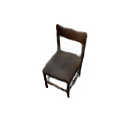} &
\includegraphics[width=1.2cm]{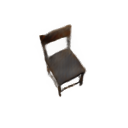}&
\includegraphics[width=1.2cm]{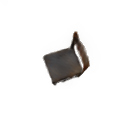}
\end{tabular}
\end{tabularx}
 \caption{Novel view synthesis of chairs from ShapeNet-V2 \cite{sitzmann2019scene} given 2 input views using our method.} 
\label{fig:chair2} 
\end{figure}

\begin{figure}
 \vspace{-5mm}
\def\tabularxcolumn#1{m{#1}}
\begin{tabularx}{\linewidth}{@{}cXX@{}}
\setlength{\tabcolsep}{0pt}
\begin{tabular}{c | ccc cc cc cc}
\multicolumn{1} {c|}{\bf Input} &  \multicolumn{9}{c}{\bf Rendered novel views}\\
\includegraphics[width=1.2cm]{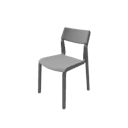} &
\includegraphics[width=1.2cm]{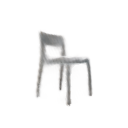} &
\includegraphics[width=1.2cm]{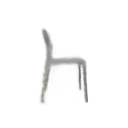} &
\includegraphics[width=1.2cm]{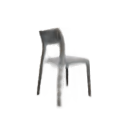} &
\includegraphics[width=1.2cm]{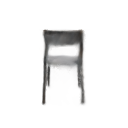} &
\includegraphics[width=1.2cm]{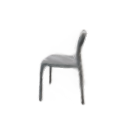} &
\includegraphics[width=1.2cm]{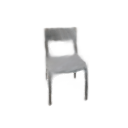} &
\includegraphics[width=1.2cm]{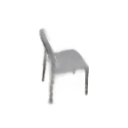} &
\includegraphics[width=1.2cm]{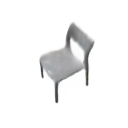}&
\includegraphics[width=1.2cm]{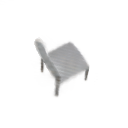}
\\
\includegraphics[width=1.2cm]{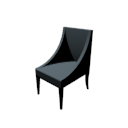} &
\includegraphics[width=1.2cm]{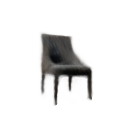} &
\includegraphics[width=1.2cm]{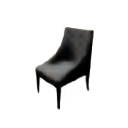} &
\includegraphics[width=1.2cm]{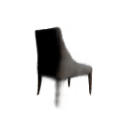} &
\includegraphics[width=1.2cm]{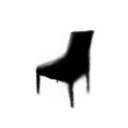} &
\includegraphics[width=1.2cm]{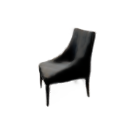} &
\includegraphics[width=1.2cm]{fig_supp/chairs_1view/4c97f421c4ea4396d8ac5d7ad0953104/0_img63_28.9155.png} &
\includegraphics[width=1.2cm]{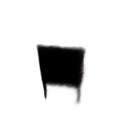} &
\includegraphics[width=1.2cm]{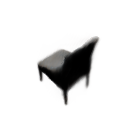} &
\includegraphics[width=1.2cm]{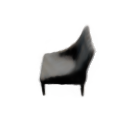} 
\\
\includegraphics[width=1.2cm]{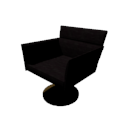} &
\includegraphics[width=1.2cm]{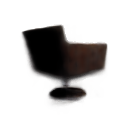} &
\includegraphics[width=1.2cm]{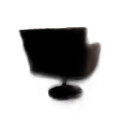} &
\includegraphics[width=1.2cm]{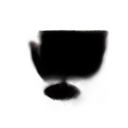} &
\includegraphics[width=1.2cm]{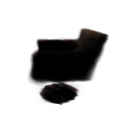} &
\includegraphics[width=1.2cm]{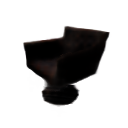}&
\includegraphics[width=1.2cm]{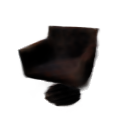}&
\includegraphics[width=1.2cm]{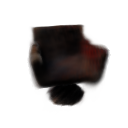} &
\includegraphics[width=1.2cm]{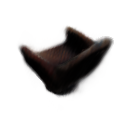} &
\includegraphics[width=1.2cm]{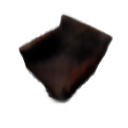} \\
\includegraphics[width=1.2cm]{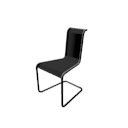} &
\includegraphics[width=1.2cm]{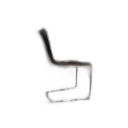} &
\includegraphics[width=1.2cm]{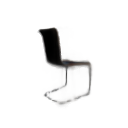} &
\includegraphics[width=1.2cm]{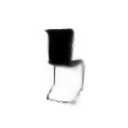} &
\includegraphics[width=1.2cm]{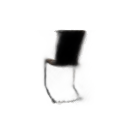} &
\includegraphics[width=1.2cm]{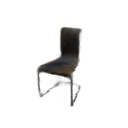}&
\includegraphics[width=1.2cm]{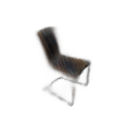}&
\includegraphics[width=1.2cm]{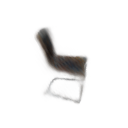} &
\includegraphics[width=1.2cm]{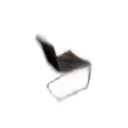} &
\includegraphics[width=1.2cm]{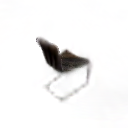} 
\end{tabular}
\end{tabularx}
 \caption{Novel view synthesis of chairs from ShapeNet-V2 \cite{sitzmann2019scene} given a single input view using our method.} 
\label{fig:chair1} 
\end{figure}

\begin{figure}
\def\tabularxcolumn#1{m{#1}}
\begin{tabularx}{\linewidth}{@{}cXX@{}}
\setlength{\tabcolsep}{0pt}
\begin{tabular}{c|c cc cc cc cc}
\multicolumn{1} {c|}{\bf Inputs} &  \multicolumn{9}{c}{\bf Rendered novel views}\\
\includegraphics[width=1.2cm]{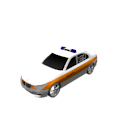} &
\includegraphics[width=1.2cm]{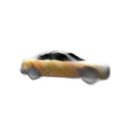} &
\includegraphics[width=1.2cm]{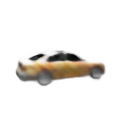} &
\includegraphics[width=1.2cm]{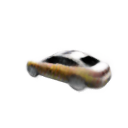} &
\includegraphics[width=1.2cm]{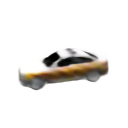} &
\includegraphics[width=1.2cm]{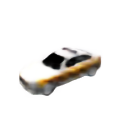} &
\includegraphics[width=1.2cm]{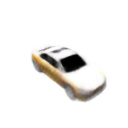} &
\includegraphics[width=1.2cm]{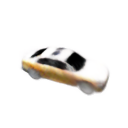} &
\includegraphics[width=1.2cm]{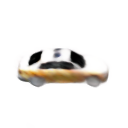}&
\includegraphics[width=1.2cm]{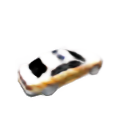}
\\
\includegraphics[width=1.2cm]{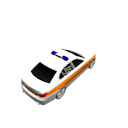} &
\includegraphics[width=1.2cm]{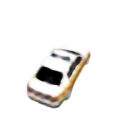} &
\includegraphics[width=1.2cm]{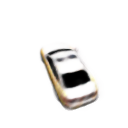} &
\includegraphics[width=1.2cm]{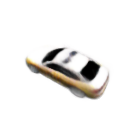} &
\includegraphics[width=1.2cm]{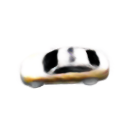} &
\includegraphics[width=1.2cm]{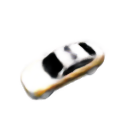} &
\includegraphics[width=1.2cm]{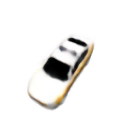} &
\includegraphics[width=1.2cm]{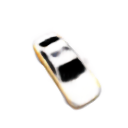} &
\includegraphics[width=1.2cm]{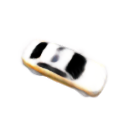}&
\includegraphics[width=1.2cm]{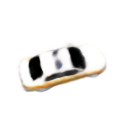} \\
\hline
\includegraphics[width=1.2cm]{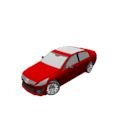} &
\includegraphics[width=1.2cm]{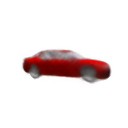} &
\includegraphics[width=1.2cm]{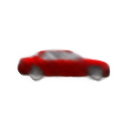} &
\includegraphics[width=1.2cm]{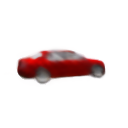} &
\includegraphics[width=1.2cm]{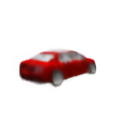} &
\includegraphics[width=1.2cm]{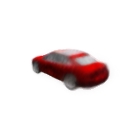} &
\includegraphics[width=1.2cm]{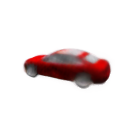} &
\includegraphics[width=1.2cm]{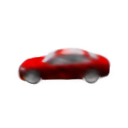} &
\includegraphics[width=1.2cm]{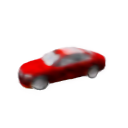}&
\includegraphics[width=1.2cm]{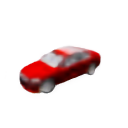}
\\
\includegraphics[width=1.2cm]{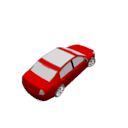} &
\includegraphics[width=1.2cm]{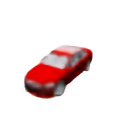} &
\includegraphics[width=1.2cm]{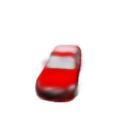} &
\includegraphics[width=1.2cm]{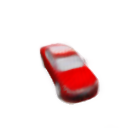} &
\includegraphics[width=1.2cm]{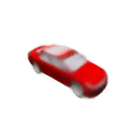} &
\includegraphics[width=1.2cm]{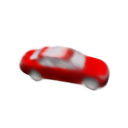} &
\includegraphics[width=1.2cm]{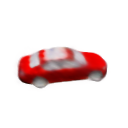} &
\includegraphics[width=1.2cm]{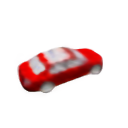} &
\includegraphics[width=1.2cm]{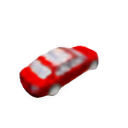}&
\includegraphics[width=1.2cm]{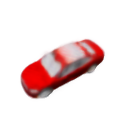} \\
\hline

\includegraphics[width=1.2cm]{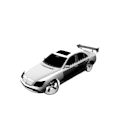} &
\includegraphics[width=1.2cm]{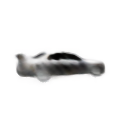} &
\includegraphics[width=1.2cm]{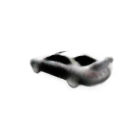} &
\includegraphics[width=1.2cm]{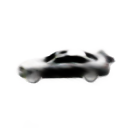} &
\includegraphics[width=1.2cm]{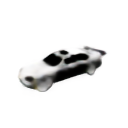} &
\includegraphics[width=1.2cm]{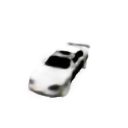} &
\includegraphics[width=1.2cm]{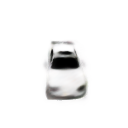} &
\includegraphics[width=1.2cm]{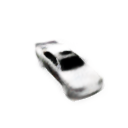} &
\includegraphics[width=1.2cm]{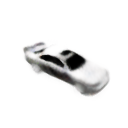}&
\includegraphics[width=1.2cm]{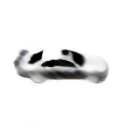}
\\
\includegraphics[width=1.2cm]{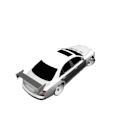} &
\includegraphics[width=1.2cm]{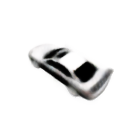} &
\includegraphics[width=1.2cm]{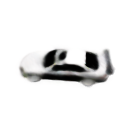} &
\includegraphics[width=1.2cm]{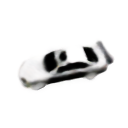} &
\includegraphics[width=1.2cm]{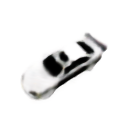} &
\includegraphics[width=1.2cm]{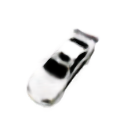} &
\includegraphics[width=1.2cm]{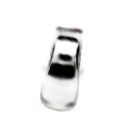} &
\includegraphics[width=1.2cm]{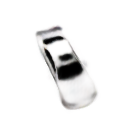} &
\includegraphics[width=1.2cm]{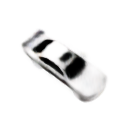}&
\includegraphics[width=1.2cm]{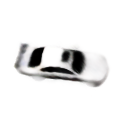} \\
\hline

\includegraphics[width=1.2cm]{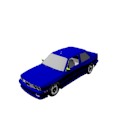} &
\includegraphics[width=1.2cm]{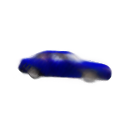} &
\includegraphics[width=1.2cm]{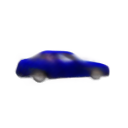} &
\includegraphics[width=1.2cm]{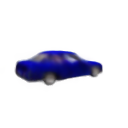} &
\includegraphics[width=1.2cm]{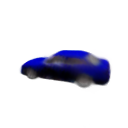} &
\includegraphics[width=1.2cm]{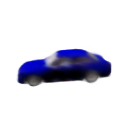} &
\includegraphics[width=1.2cm]{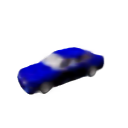} &
\includegraphics[width=1.2cm]{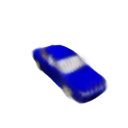} &
\includegraphics[width=1.2cm]{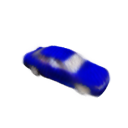}&
\includegraphics[width=1.2cm]{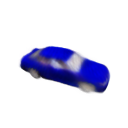}
\\
\includegraphics[width=1.2cm]{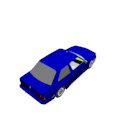} &
\includegraphics[width=1.2cm]{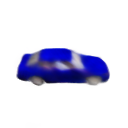}&
\includegraphics[width=1.2cm]{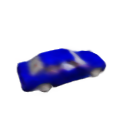} &
\includegraphics[width=1.2cm]{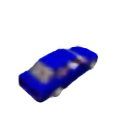} &
\includegraphics[width=1.2cm]{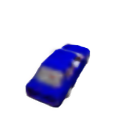} &
\includegraphics[width=1.2cm]{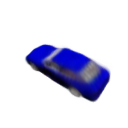} &
\includegraphics[width=1.2cm]{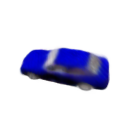} &
\includegraphics[width=1.2cm]{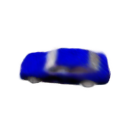} &
\includegraphics[width=1.2cm]{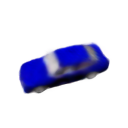} &
\includegraphics[width=1.2cm]{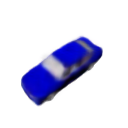}
\end{tabular}
\end{tabularx}
 \caption{Novel view synthesis of cars from ShapeNet-V2 \cite{sitzmann2019scene} given 2 input views using our method.} 
\label{fig:car2} 
\end{figure}

\begin{figure}
 \vspace{-5mm}
\def\tabularxcolumn#1{m{#1}}
\begin{tabularx}{\linewidth}{@{}cXX@{}}
\setlength{\tabcolsep}{0pt}
\begin{tabular}{c | ccc cc cc cc}
\multicolumn{1} {c|}{\bf Input} &  \multicolumn{9}{c}{\bf Rendered novel views}\\
\includegraphics[width=1.2cm]{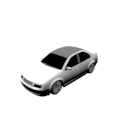} &
\includegraphics[width=1.2cm]{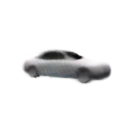} &
\includegraphics[width=1.2cm]{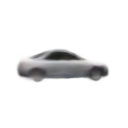} &
\includegraphics[width=1.2cm]{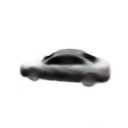} &
\includegraphics[width=1.2cm]{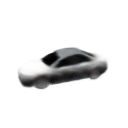} &
\includegraphics[width=1.2cm]{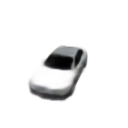}&
\includegraphics[width=1.2cm]{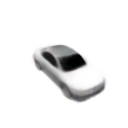}&
\includegraphics[width=1.2cm]{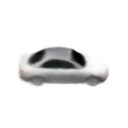} &
\includegraphics[width=1.2cm]{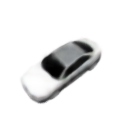} &
\includegraphics[width=1.2cm]{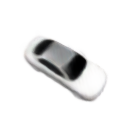} \\
\includegraphics[width=1.2cm]{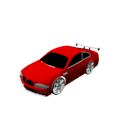} &
\includegraphics[width=1.2cm]{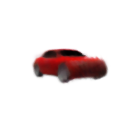} &
\includegraphics[width=1.2cm]{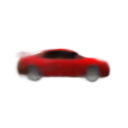} &
\includegraphics[width=1.2cm]{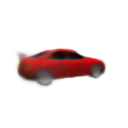} &
\includegraphics[width=1.2cm]{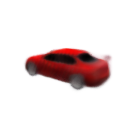} &
\includegraphics[width=1.2cm]{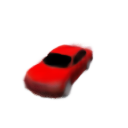}&
\includegraphics[width=1.2cm]{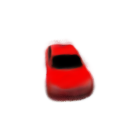}&
\includegraphics[width=1.2cm]{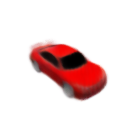} &
\includegraphics[width=1.2cm]{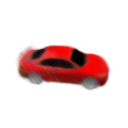} &
\includegraphics[width=1.2cm]{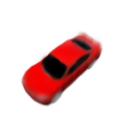}\\
\includegraphics[width=1.2cm]{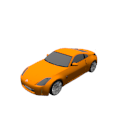} &
\includegraphics[width=1.2cm]{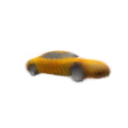}&
\includegraphics[width=1.2cm]{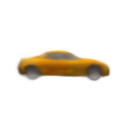} &
\includegraphics[width=1.2cm]{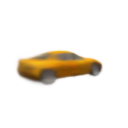} &
\includegraphics[width=1.2cm]{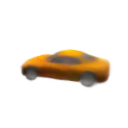} &
\includegraphics[width=1.2cm]{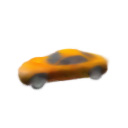} &
\includegraphics[width=1.2cm]{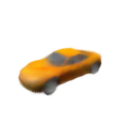} &
\includegraphics[width=1.2cm]{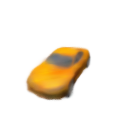} &
\includegraphics[width=1.2cm]{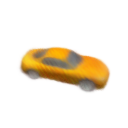} &
\includegraphics[width=1.2cm]{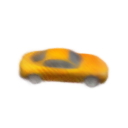}\\
\includegraphics[width=1.2cm]{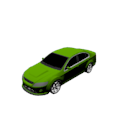} &
\includegraphics[width=1.2cm]{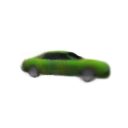} &
\includegraphics[width=1.2cm]{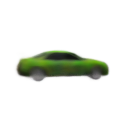} &
\includegraphics[width=1.2cm]{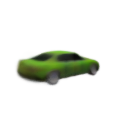} &
\includegraphics[width=1.2cm]{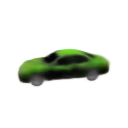} &
\includegraphics[width=1.2cm]{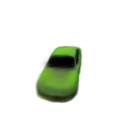} &
\includegraphics[width=1.2cm]{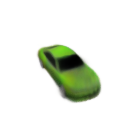} &
\includegraphics[width=1.2cm]{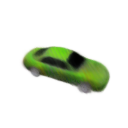} &
\includegraphics[width=1.2cm]{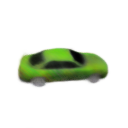} &
\includegraphics[width=1.2cm]{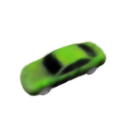} 

\end{tabular}
\end{tabularx}
 \caption{Novel view synthesis of cars from ShapeNet-V2 \cite{sitzmann2019scene} given a single input view using our method.} 
\label{fig:car1} 
 \vspace{-5mm}
\end{figure}

\begin{figure}
\centering
 \vspace{-50mm}
\def\tabularxcolumn#1{m{#1}}
\begin{tabularx}{\linewidth}{ccXXc}
\setlength{\tabcolsep}{0pt}
\begin{tabular}[p]{c cc cc cc cc c}
\multicolumn{1}{c}{\bf Input} &  \multicolumn{9}{c}{\bf Rendered novel views}\\
&
\rotatebox{90}{\quad LFN} & \includegraphics[width=1.3cm]{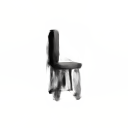} &
\includegraphics[width=1.3cm]{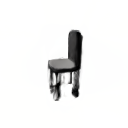} &
\includegraphics[width=1.3cm]{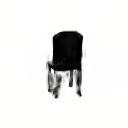} &
\includegraphics[width=1.3cm]{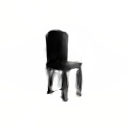} &
\includegraphics[width=1.3cm]{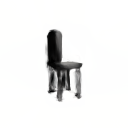} &
\includegraphics[width=1.3cm]{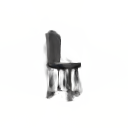} &
\includegraphics[width=1.3cm]{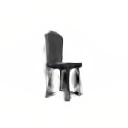} &
\includegraphics[width=1.3cm]{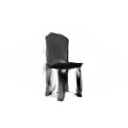} \\
\includegraphics[width=1.3cm]{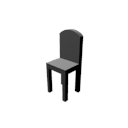}&
\rotatebox{90}{\quad Ours} & \includegraphics[width=1.3cm]{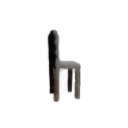} &
\includegraphics[width=1.3cm]{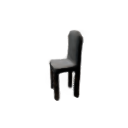} &
\includegraphics[width=1.3cm]{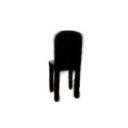} &
\includegraphics[width=1.3cm]{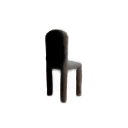} &
\includegraphics[width=1.3cm]{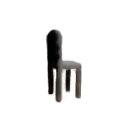} &
\includegraphics[width=1.3cm]{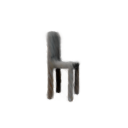} &
\includegraphics[width=1.3cm]{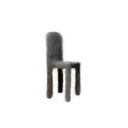} &
\includegraphics[width=1.3cm]{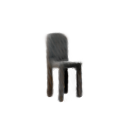}  \\
&
\rotatebox{90}{\quad GT} &
\includegraphics[width=1.3cm]{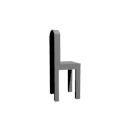} &
\includegraphics[width=1.3cm]{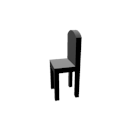} &
\includegraphics[width=1.3cm]{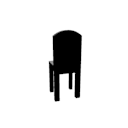} &
\includegraphics[width=1.3cm]{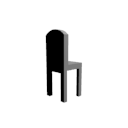} &
\includegraphics[width=1.3cm]{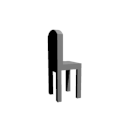} &
\includegraphics[width=1.3cm]{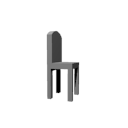} &
\includegraphics[width=1.3cm]{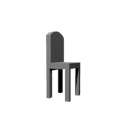} &
\includegraphics[width=1.3cm]{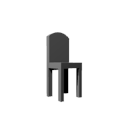} \\
\hline
&

\rotatebox{90}{\quad LFN} &
\includegraphics[width=1.3cm]{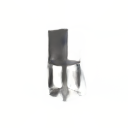} &
\includegraphics[width=1.3cm]{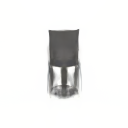} &
\includegraphics[width=1.3cm]{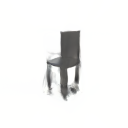} &
\includegraphics[width=1.3cm]{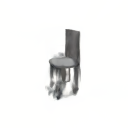} &
\includegraphics[width=1.3cm]{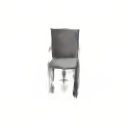} &
\includegraphics[width=1.3cm]{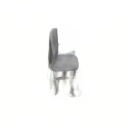} &
\includegraphics[width=1.3cm]{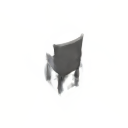} &
\includegraphics[width=1.3cm]{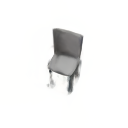} 
\\
\includegraphics[width=1.3cm]{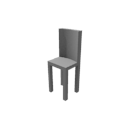}&
\rotatebox{90}{\quad Ours} &
\includegraphics[width=1.3cm]{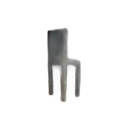} &
\includegraphics[width=1.3cm]{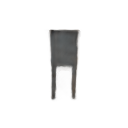} &
\includegraphics[width=1.3cm]{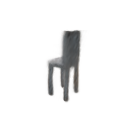} &
\includegraphics[width=1.3cm]{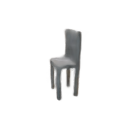} &
\includegraphics[width=1.3cm]{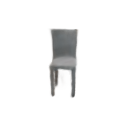} &
\includegraphics[width=1.3cm]{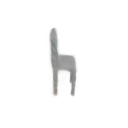} &
\includegraphics[width=1.3cm]{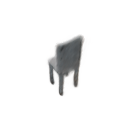} &
\includegraphics[width=1.3cm]{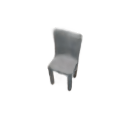}  \\
&

\rotatebox{90}{\quad GT} &
\includegraphics[width=1.3cm]{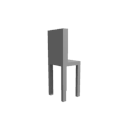} &
\includegraphics[width=1.3cm]{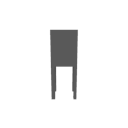} &
\includegraphics[width=1.3cm]{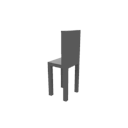} &
\includegraphics[width=1.3cm]{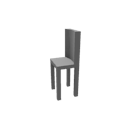} &
\includegraphics[width=1.3cm]{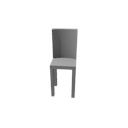} &
\includegraphics[width=1.3cm]{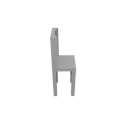} &
\includegraphics[width=1.3cm]{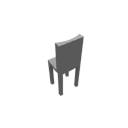} &
\includegraphics[width=1.3cm]{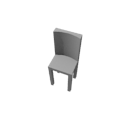}\\
\hline
&
\rotatebox{90}{\quad LFN} &
\includegraphics[width=1.3cm]{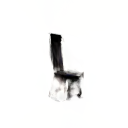} & 
\includegraphics[width=1.3cm]{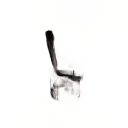} &
\includegraphics[width=1.3cm]{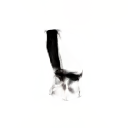} &
\includegraphics[width=1.3cm]{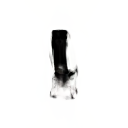} &
\includegraphics[width=1.3cm]{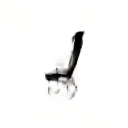} &
\includegraphics[width=1.3cm]{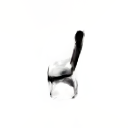} &
\includegraphics[width=1.3cm]{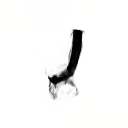} &
\includegraphics[width=1.3cm]{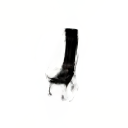} 
\\
\includegraphics[width=1.3cm]{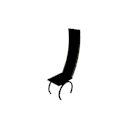}&
\rotatebox{90}{\quad Ours} &
\includegraphics[width=1.3cm]{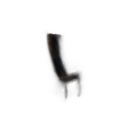} &
\includegraphics[width=1.3cm]{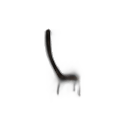} &
\includegraphics[width=1.3cm]{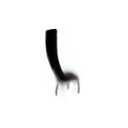} &
\includegraphics[width=1.3cm]{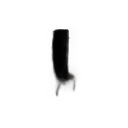} &
\includegraphics[width=1.3cm]{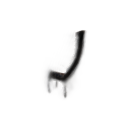} &
\includegraphics[width=1.3cm]{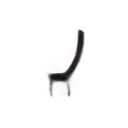} &
\includegraphics[width=1.3cm]{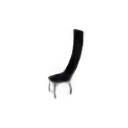} &
\includegraphics[width=1.3cm]{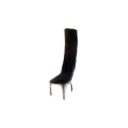} 
\\
&
\rotatebox{90}{\quad GT} &
\includegraphics[width=1.3cm]{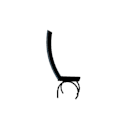} &
\includegraphics[width=1.3cm]{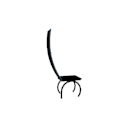} &
\includegraphics[width=1.3cm]{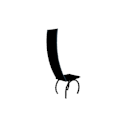} &
\includegraphics[width=1.3cm]{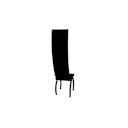} &
\includegraphics[width=1.3cm]{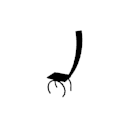} &
\includegraphics[width=1.3cm]{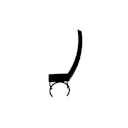} &
\includegraphics[width=1.3cm]{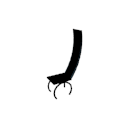} &
\includegraphics[width=1.3cm]{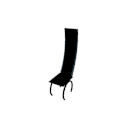}\\ 
\end{tabular}
\end{tabularx}
 \caption{Qualitative comparison to \cite{sitzmann2021light} on novel view synthesis of chairs from ShapeNet-V2 \cite{sitzmann2019scene} using a single input view.} 
\label{fig:chairs-lfn} 
 \vspace{-5mm}
\end{figure}


\end{document}